# A Formal Definition and Meta-model for a Machine Theory of Mind

Fabio Cuzzolin
Institiure for AI, Data Analysis and Systems (AIDAS)
School of Engineering, Computing and Mathematics
Oxford Brookes University

## Abstract

*This paper proposes, for the first time, a rigorous formal definition of the concept of Machine Theory of Mind, based on principles supported by evidence from cognitive psychology, neuroscience and artificial intelligence, and uses the above as a lens to examine state-of-the-art and current efforts in the field, driving a potential agenda for further research there able to "crack" the problem. It also advances a general holistic meta-model for Maching Theory of Mind, and examines the state of the art when it comes to empirically benchmarking such models.*

# Main Paper

## Preface

## 1 Introduction

### 1.1 Theory of mind

#### 1.1.1 Concept from cognitive psychology

**Concept**

Theory of mind (ToM) is the problem of dynamically learning to understand the thinking process of an agent or class of agents external to ourselves. It can also be seen as the ability of the human mind to attribute mental states to others and is a key component of human cognition. In order to understand other people's mental states or viewpoint and to have successful interactions with others within social and occupational environments, this form of social cognition is arguably essential.

**ToM approches**

One of the three dominant theoretical approaches to ToM is called *Theory-Theory* (TT). This maintains that the knowledge concerning mental aspects lies in some structures whose contents are represented in the form of theories, and that the development of a primitive understanding of mental states in children comes with a radical conceptual change[1,2,3]. Consistently with this idea (the *child scientist*), children often elaborate theories and formulate hypotheses in a propositional form, which they try to confirm or disprove through experience, just as a scientist would do[4,5]. The TT assumption that an individual generates, and stores, a very large number of theories about other people and their behaviour is problematic, as such an approach would be cognitively wasteful.

For some, TT means having an (innate) structure of the human mind, which matures over time. In this sense, having TT is equivalent to choosing  class of cognitive models (e.g., *belief-desire-intention*, BDI)[6]. Programming languages for BDI exist, e.g., Jack[7], Jadex[8] and 2APL[9].

*Simulation-Theory* (ST)[10], instead, states that an individual uses (often automatically and unconsciously) a mechanism that simulates other people's mental states using his own reasoning processes. In other words, said individual activates processes that try to provide an answer to the question “How would I behave if I were in their shoes, having their own beliefs and desires?”. If we assume that everyone's mind works in a similar way, it becomes possible to predict someone else's behaviour by predicting how we ourselves would behave in the same situation[11].

A third approach is provided by the set of *modular theories*[12], which maintain that there is a module specialised for ToM in the brain. In this module, all the mechanisms and theories that allow an individual to attribute mental states to others and to predict someone else's behaviour are specified. This module could be the product of an evolutionary process, since the ability to accurately predict others' behaviour in the past must have been crucial to guarantee the survival of the species. These ideas would explain why Theory of Mind seems to develop uniformly among different peoples and cultures.

The need to overcome the dichotomy between TT and ST has been highlighted by several authors[1314].

**Neuroscience of ToM**

The neuroscience of Theory of Mind is an active subject of studies[15], leading to independent definitions of the concept[16]. A lot of work has focused on *mirror neurons*[17], as potential candidate structures for human ToM in the brain. Studies[18] have shown that empathy-related processing of emotional facial expressions recruits brain areas involved in mirror neuron and Theory-of-Mind mechanisms. Despite a declining interest in them in the past 10-15 years[19], research on their origin has confirmed their importance for domain-general visual-motor associative learning. However, brain stimulation suggests that mirror-neuron brain areas contribute to low-level processing of observed actions (e.g., distinguishing types of grip) but not to high-level action interpretation[20]. Whether there are specialised ToM structures in the brain is still debated[21]. A rather outdated review[22] suggests that several distinct brain regions may be activated during ToM reasoning, forming an integrated functional "network." The imaging findings suggest that there are several "core" regions in the network—including parts of the prefrontal cortex and superior temporal sulcus— while several more "peripheral" regions may contribute to ToM reasoning in a manner contingent on relatively minor aspects of the ToM task. More recent research[23] indicates that a degree of variability of the activation during indirect communication of "ToM areas" such as the medial prefrontal cortex and bilateral temporo-parietal junction. Other studies on indirectness of communication raise questions about the role of different communicative function types in drawing upon those regions.

Overall, neuroscience does not appear to be (yet) a position to strongly inform the design of a Machine Theory of Mind, alternative to the (re-)application of general-purpose, not specifically trained/designed large language models to the task. Still, it can provide valuable evidence, while further advances in the field are very likely.

## 1.2 Machine theory of mind

### 1.2.1 Motivation and interdisciplinary appeal

The motivation for having a Machine Theory of Mind (MToM) is strong[24], as the concept, being related a fundamental tenet of intelligence, appeals to many different scientific communities: artificial intelligence[25] and its applications[26], cognitive, experimental and developmental psychology[27,28,29] (where the concept originates), neuroscience[30,31,32], but also philosophy[33], ethics[34,35], anthropology[36], linguistics[37,38], social sciences[39], ethology[40,41], robotics[42], human-computer interaction[43], healthcare[44] and many others.

Within AI, many believe this an essential steps towards human-level or superior collective intelligence[45,46,47]. Most recently, for instance, the role of ToM in agentic AI is starting to draw attention[48]. The topic is of interets to the philosophical community as well[49,50]. Its implications for robotics are investigated in[51], but also in[52], from a visual behaviour point of view; the interest within the robotics community is, in fact, long-standing[53]. Latent Theory of Mind, a decentralized diffusion policy architecture allowing multiple manipulators with their own perception and computation to collaborate with each other towards a common task goal with or without explicit communication is proposed in[54]. More recently, the application of the concept to autonomous driving (e.g., for trajectory prediction) has been explored[55], including its implications for the explainability of robot cars' behaviour[56] or safety[57]. A number of startup companies exploring ToM for AVs already exists, e.g. Perceptive Automata, Motor AI[1] and ISEE AI[2].

### 1.2.2 Recent surveys

Several attempts have been made at endowing machines with a theory of mind, following a number of diverse, and rather disconnected research strands, with most work focussing on learning of preferences. A good recent survey on the relevant literature can be found in[58]. Another, more recent review is[59]. A recent categorisation of approaches to Machine ToM can be found in[60], distinguishing in particular betweem *cognitive-based and black-box algorithms* and *bio-inspired models*, as well as considering applications. Computational theory of mind is the focus of recent book by Colombo and Piccinini[61].

### 1.2.3 Recent efforts

Recent exiting work has been done since those surveys, exploring a variety of different concepts and approaches.

---

[1] https://www.motor-ai.com/

[2] https://engineventures.com/companies/isee

Among the most interesting approaches are *symmetric* machine theory of mind[62], a multi-agent, reinforcement learning environment in which agents gaining ToM boost their chance of success. Multi-agent collaboration via ToM is also the focus of[63], in which LLM-based agents are evaluated in a multi-agent cooperative text game and compared with Multi-Agent Reinforcement Learning (MARL)[64] and planning-based baselines, showing evidence of emergent collaborative behaviors and high-order Theory of Mind capabilities among LLM-based agents. An interesing "meta-mind" approach is proposed in[65], in which agents learn not only to predict and plan over their own beliefs, but also to inversely reason goals and beliefs from their own behavior trajectories. *Scaffolding* approaches to AI agents are also being actively pursued[66]. Theory of mind is a means to explain agent behaviour via root cause analysis in[67]. Its effect on improving trust in humans is considered in[68]. COKE[69] is a cognitive knowledge graph for machine theory of mind, which formalizes ToM as a collection of 45k+ manually verified cognitive chains that characterize human mental activities. Multimodal ToM question-answering[70] uses scalable Bayesian inverse planning with language models. The paper also introduces a rare multimodal benchmark called MMToM-QA, which evaluates machine ToM both on multimodal data and on different kinds of unimodal data about a person's activity in a household environment. Adversarial approaches are explored in various papers[71], while a recent game-theoretical effort can be found in[72]. A memory-augmented theory of mind network called TOMMY is advanced in[73]. Multimodal multi-agent ToM is also core to the recent MuMA-ToM model[74].

In[75], a MtoMnet network encoding contextual cues (scene videos and object locations) and integrating them with person-specific cues (human gaze and body language) in a separate MindNet for each person is proposed in three variants: two involving fusion of latent representations and one involving re-ranking of classification scores.

An interesting effort to bridge ML (machine learning) and HCI (human-computer interaction) aspects of "mutual" ToM is[76], where the authors argue that uni-directional, first-order models (e.g., a human's mental model of an AI agent) are not enough to achieve Machine ToM; rather, at least second-order models are needed.

Interpretability is studied in[77]. TOMA[78] provides a mechanism for computational theory of mind based on abstractions of single beliefs into higher-level concepts, triggered by social norms and roles. Decompose-ToM[79] tries to enhance ToM reasoning in LLMs by recursively simulating user perspectives and decomposing the ToM task into a simpler set of tasks: subject identification, question-reframing, world model updation, and knowledge availability.

The role of machine ToM in social intelligence is discussed in[80,81]. Social reasoning from the point of view of LLMs is considered in[82] over six different tasks: the authors find that their behaviour is far from robust, struggling in particular with adversarial examples. Planning, instead, is the focus of[83].

EnigmaToM[84] is an original a novel neuro-symbolic framework that enhances ToM reasoning by integrating a neural knowledge base of entity states (*Enigma*) for a psychology-inspired iterative masking mechanism that facilitates accurate perspective-taking, and knowledge injection that elicits key entity information. Enigma generates structured knowledge of entity states to build spatial scene graphs for belief tracking across various ToM orders and enrich events with fine-grained entity state details.

### 1.2.4 Large language models and ToM

Many recent studies have analysed the (real of supposed) ToM capabilities of *large language models* (LLMs)[85,86,87,88,89], mostly following an empirical approaches using experiments[90], rather than from principles. Ullman[91] have shown that Large Language Models fail on trivial alterations to Theory-of-Mind tasks[92]. Ma et al.[93] have proposed a taxonomy articulated into 7 mental states categories, as benchmark for LLM capabilities. Strachan et al.[94] have advanced a comprehensive battery of measurements that aim to measure different Theory of Mind abilities, from understanding false beliefs to interpreting indirect requests and recognizing irony and faux pas, and tested both humans and language models on it, finding that "faux pas" was the only test where LLaMA2 outperformed humans. The challenges faced by deep learning approaches to ToM are considered in[95]; the authors conclude by encouraging researchers to investigate Theory of Mind in complex open-ended environments to avoid shortcuts (a typical issue within, for instance, reinforcement learning agents). The robustness of ToM in LLMs is also discussed in[96], where the authors introduce a novel dataset of 68 tasks for probing ToM in LLMs, including potentially challenging variations which are assigned to 10 complexity classes. An interesting idea can be found in Tang and Belle's paper[97], where ToM in LLMs is delegated to an external symbolic executor. The role of prompting is explored in[98], with self-presentation being the focus of[99].

It is fair to say most authors come to skeptic conclusions about the supposed ToM abilities of large language models. GPT-4o is found wanting in[100]. The analysis is extended to Foundation Models in[101]. The interaction between ToM and the rising field of Agentic AI is explored in[102], where LLMs are framed as agents and their reasoning is guided using psychology-inspired prompt functions, termed "cognitive tools".

### 1.2.5 Uncertainty and machine ToM

*Bayesian theory of mind* is so far the mainstream approach to uncertainty modelling in ToM[103,104,105]. A rather original take on quantifying uncertainty in Machine ToM over time is found in[106], where Bayesian inference is used to maintain a posterior belief about mental states as a probability distribution, updated with observational data. Interestingly, a definition of ToM complexity via *Discrete World Models*, in terms of the number of states necessary to solve a task correctly, is advanced in[107]. A hierarchical Theory of Mind Model based on strategy inference, called BELIEFS, is proposed in[108], as a probabilistic ToM framework that infers latent level-0 strategies directly from behavior using a Hidden Markov Model. The implications of unpredictability for anthropomorphism in Machine ToM is touched upon, quite originally, in[109]. Multimodal ToM using weak-to-strong Bayesian reasoning is proposed in[110].

### 1.2.6 AGI and machine ToM

The topic is obviously related to Artificial General Intelligence (AGI)[111], concept still in search for a proper formal definition[112,113]. Schneider maintains ToM is a core component of AGI[114]. The implications of Machine ToM for "caring machines" are discussed in another interesting recent paper from a marketing perspective[115]. Whether language models truly "understand" us[116] is the question asked in[117].

### 1.2.7 Remarks

Most or all past efforts have arguably been just partial attempts, because of an a-priori choice of a particular technique and a general lack of a formal definition of the concept of Machine Theory of Mind. Basically, no efforts have been made to rigorously define the concept before attempting to tackle it. A recent cursory attempt at a general characterisation of the problem, however, can be found in[118].

Relevantly to this contribution, ontological guidelines for a mutual Human-AI ToM are proposed in[119], where a *Perception Theory* (PT) approach, opposed to both ST and TT, is brougth forward. PT posits that we directly perceive the mental states of others. From a philosophical stance opposed to metaphysical realism, the authors argue for a ToM that includes negotiation of reference between agents, emergence of ToM from social interactions (rather than pre-definition), a focus on non-conceptual intentions, and flexible representations of objects, thereby questioning the notion of ground truth in AI. Some elements of this, albeit from a realist point of view, are part of our own meta-model.

## 1.3 Rationale: prediction at the "tail"

Theory of Mind has been evolutionary developed to boost humans' success in their evolving environment. To mimic humans' success, especially in a world made by humans for humans, it is crucial for machines to understand the other intelligences sharing the environment with them.

This implies being able to predict their future behaviour and intentions, but also to explain the why of their behaviour. This is a substantial difference with pattern recognition-based approaches[120,121], including, for instance, trajectory prediction[122], which cannot explain the reasoning behind their predictions. This is arguably the most important rationale for a machine needing a Theory of Mind. In a way, this can be seen as the inverse of AI explainability[123]: it is about explaining humans to machines.

The second reason comes from allowing more accuracy predictions and better interactions with the environment. Prediction-wise, ToM alone has the potential to cope with "long tail" (rare) situations. People, in particular, have rich mental lives, whose potential is only partially realised in their observable behaviour. This is still an open research question: by which mechanisms Theory of Mind is able (or not) to overcome the lack of observational evidence, if itself is built upon continuous observation in a life-long learning process? Theory-theory can potentially address this, but, unless "theories" are somehow innate to the human brain (or, in a machine's case, given to it as prior, symbolic knowledge), one may argue theories are also inferred from evidence.

Both of these are strong argument in support of the usefulness of ToM for AI alignment[124,125,126,127] (including moral judgements[128]) and safety[129], both a source of great and growing concern in the AI community and society at large.

### 1.4 Contributions

The main contributions of this paper are:

- A world-first attempt to enunciate the Principles of a Machine Theory of Mind, including its expected outputs, which we call "insights".
- Again, for the first time, the proposition of a formal definition of the concept general enough to encompass all the expected features and adhering to the above Principles.
- A critical review the state of the art through the lenses of the above Principles and formal definition, in order to identify research gaps and shortcomings and guide future research in this area.
- An outline of a holistic meta-model for a Machine Theory of Mind and an analysis of its main components.
- Based on the above meta-model, a proposed, tentative agenda for future work in the field.

## 2 Principles of a machine theory of mind

Indeed, despite the consideral amount of work conducted, the field has so far arguably failed to provide a formal definition of the concept of a "Machine Theory of Mind". As a result, scientists have been rather "shooting in the dark", proposing (partial) models capable of addressing only aspects of the problem, as partial, implicit operative definitions of the concept. As stated above, here we depart from all existing efforts to first and foremost formulate a set of Principles a Machine Theory of Mind should adhere to, to then propose a rigorous formal definition for the problem. Such principles cannot be arbitrary axioms, but need to be supported by the body of evidence the three intertwined fields of AI, cognitive psychology and neuroscience can provide.

In turn, this will be used as a lens to interpret the progress made so far, to guide us towards a "meta-model" for machine theory of mind, identify outstanding research gaps and outline a pathway for the future.

### 2.1 Theory of mind as a lifelong learning process

There is strong evidence from psychology[130] (in particular for autistic children[131]) but also AI[132,133], that ToM is really a lifelong learning process[134]. ToM abilities in humans are built over a number of years, as a result of multiple varied interactions with other people, and appear in children at around 3-4 years of age[135]. This is indicative, on the one hand, of the sheer amount of data an entity needs to gather to achieve ToM; on the other, it tells us that any approaches following a traditional "one-off" learning strategy are quite doomed to fail. Overall, the role of learning and observation[136] seems to be crucial[137,138] and should not be left out in any attempt to model a Machine ToM. In fact, awareness of this requirement is finally rising within AI[139,140].

Very relevantly to this argument, a recent paper[141] also stresses the fact that existing benchmarks overlook the temporal evolution that characterizes real-world social interactions. Its proposed DynToM benchmark, in opposition, is specifically designed to evaluate LLMs' ability to understand and track the temporal progression of mental states across interconnected scenarios.

This leads to our first Principle:

***Principle 1**: ToM models should be continually updated based on experience.*

## 2.2 Role of the simulation of the self

What kick-starts this lifelong learning experience? Related to that, what model can we use for agents we have never interacted with before? It is reasonable to presume, as supported by the literature, that the starting point is the simulation of the self[142,143], as a child[144] applies their own reasoning process to other people while interacting with them[145]. This is supported by research on the role of attribution[146]. Interesting, in this sense, is the recent COMMON-TOM banchmark which attempts to establishe a common ground between my own view and those of others[147].

***Principle 2**: In the absence of evidence, Machine Theory of Mind should leverage an agent's own thinking process to simulate others.*

## 2.3 Learning from observation and interaction

Humans interact with their environment (including other sentient agents) in two ways: by observing it (including being explained how to do things, for instance by a parent or a teacher), or by acting within it. Few studies look at the importance of these two learning mechanisms for Theory of Mind formation[148], mainly from a reinforcement learning stance[149], but neglecting the pure observation element which is crucial to develop new "theories". Communication with other agents[150] is also a form of interaction with the environment. Nevertheless, it is reasonable to argue that a Machine Theory of Mind should tap *both* sources, just as human do. In machine learning, those roughly correspond to *supervised learning*[151] and *reinforcement learning*[152,153], respectively.

Supervised learning (SL) can be employed to achieve ToM abilities by assessing the difference between the observed and ToM-predicted behaviour of the other agent. It is harder, if not outright impossible, to imagine machine ToM models trained on "true" mental state information provided by humans – in all realistic scenarios such information is unavailable (and possibly non-existent, given its hidden nature), making this kind of supervision appear artificial. Humans also do not have access to the mental states of the others in their lifelong learning experience, unless directly communicated by the other, which seldom happens[154,155], albeit experiments in this sense in, e.g., robotics exist.

Reinforcement learning[156] (RL), on the other hand, is a sequential learning framework[157] in which the acts of an agent change the state of the world it belongs to. The goal is for the agent to learn the "policy" (i.e., series of future actions) which it expects to maximise its cumulative future reward[158]. Reinforcement learning can be used to continually refine ToM abilities[159] as the agent keeps interacting with an environment containing other intelligent agents, as follows: the agent uses its ToM abilities to predict the environment's future evolution; it acts (partly) based on those predictions; it gets rewarded or penalised as a result of its actions; it eventually uses this feedback to improve its model of the other agents' thinking.

The two learning processes are *interlaced*: however, sometimes we observe without acting at all, and can still learn in the process[160] (e.g., by watching a movie or reading a story, or observing what happens to our friends). Notice that the speed of the interaction can be a factor (see Sec. 2.1): even when we do learn/update our theory of other minds by interacting with the environment and other agents, sometimes we might not have time to verify our predictions before taking the next decision.

> ***Principle 3****: A Machine Theory of Mind should be learned from both observing and interacting with the environment, depending on the situation and the dynamics of the environment itself.*

## 2.4 Theory of mind as both theory and simulation

There is evidence in support of the fact that ToM arguably exploits both a simulation approach[161] (as evidenced in the mirror neurons structure of our brain[162]) and a theory approach[163]. Whenever there is insufficient information to drive an adequate simulation of the other, hypotheses on the other's mental states are produced and new data is acquired to validate them[164]. This implies the existence of an *active learning* component in which the ToM agent acts to gather more information to disambiguate the different hypotheses one makes about the other's intentions or mental states. For instance, if I see a friend tap nervously on the table, I might be unsure whether they are bored with my conversation or are worrying about something: I can then ask them (cautious) questions to find out. This seems to imply some sort of statistical hypothesis testing[165,166] based on probabilistic modelling[167]. In this sense, then, the TT component of theory of mind also relates to the issue of modelling uncertainty (Section 2.7 and Principle 8).

The following two principles thus follow:

> ***Principle 4****: Machine ToM models should follow a hybrid approach in which both the simulation of others' mental states and the formulation and validation of hypotheses have a role, combining both mainstream approaches to Theory of Mind.*

> ***Principle 5****: Machine Theory of Mind should include an active learning component in which new evidence is sought to resolve ambiguities.*

Active learning is very much related to uncertainty quantification (see Sec. 2.7), as the perceived level of uncertainty can be used to drive the acquisition of new data or information[168]. More in general, people may focus their attention on certain aspects of the environment rather than others, in order to more efficiently generate hypotheses on the other's mental state. This active learning mechanism can be seen as an "action" in a reinforcement learning setting, where, however, the purpose is to gather more information to validate hypotheses rather than to achieve a reward.

## 2.5 Depths of mentalisation

The main characteristic of ToM is the depth of recursion: this is termed *higher-order theory of mind*[169], also known as *Depth of Mentalising* (DoM)[170] (e.g., "I think that you think that I think"). The emergence of higher-order recursive ToM skills in children, however, has been much less studied than the emergence of first-order ToM[171]. Still, higher-order theory of mind is especially useful in unpredictable negotiations[172], negotiations with incomplete information[173], and games of limited bidding[174].

This is dealt with in different ways within the different streams of Machine Theory of Mind research, if at all. Given its consequences, it is fair to maintain that:

> ***Principle 6***: *Machine Theory of Mind should be capable of handling higher-order ToM.*

## 2.6 Outputs of theory of mind

As we enunciate the Principles a Machine Theory of Mind should obey, inspired from the way ToM works in humans, we need to ask the next fundamental question: what output(s) does an MToM framework generate as a process? "The thinking process of the other", an expression that naturally comes to mind, is a rather generic term.

Consider a number of scenarios:

- Children walking on the pavement spot an ice cream van[175] on the other side of the road: are they going to cross to get the ice cream? Here the objective is to predict an agent's *future behaviour* (*intention* that *materialises*).
- A person pulls a strange face while drinking coffee: *why* are they doing that? The objective, in this second example, is *explaining* the other's observed behaviour. In turn, this explanation can assume multiple forms:
    - What are they *thinking* (in terms of rational thoughts)?
    - What are they *feeling* (e.g., pleasure, disgust)?
    - What is their *emotional state* (e.g., anger, distress)?
- A person is running in the street with a knife: what is their *intention*? (are they going to kill somebody?).

It is clear that by "Theory of Mind" we intend the ability to generate *an entire range of outputs* (which we term here *insights*), most of them corresponding to mental states, others observable, without any one of them being necessarily seen as "the" preferential output, depending on the scenario.

> ***Principle 7***: *(Machine) Theory of Mind is a multi-task problem[1], in which manifold outputs of a different nature are sought.*

From an AI point of view, formalising such a multi-task problem implies deciding the mathematical form of the outputs the model will generate[176] (e.g., a vector, a set, as in conformal learning[177], a tensor[178], a probability distribution[179], a credal set[180,181,182], a random set[183,184]). In this paper, for the sake of generality, we do not make assumptions about the nature of the insights sought, or on the mathematical form of their representation.

## 2.7 Influence of uncertainty

By definition, theory of mind is subject to heavy uncertainty[185,186], as the mental states of others are, by definition, not directly observable[187,188], and human behaviour is often very unpredictable[189].

> ***Principle 8***: *The Theory of Mind process is intrinsically uncertain, as mental states and other features of another agent's mind are by definition not directly observable.*

Any realistic and effective machine ToM should then model the relation between the quantity and quality of the data at the observer's disposal and the uncertainty[190] affecting its predictions. This uncertainty is *epistemic* in nature[191,192,193], as can be reduced by acquiring new data and/or information[194], a concept clearly on a rise within machine learning[195].

# 3 Formalising machine Theory of Mind

## 3.1 Formal definition

Let us introduce the following notation to allow us to propose a world-first formal definition of machine ToM.

| | | |
|---|---|---|
| $\omega$ | → | The *Observer*: the (Machine) intelligence endowed with Theory of Mind); |
| $a$ | → | Any (other) *Agent*; |
| $\mathcal{A}$ | → | The collection of all possible (other) Agents; |
| $A \subseteq \mathcal{A}$ | → | A class of Agents; |
| $e$ | → | The *Environment* as observed by the Observer $\omega$; |
| $\mathcal{E}$ | → | The space of possible Environment configurations; |
| $y$ | → | (Series of) Observations made within an Environment $e$; |
| $m$ | → | A (ToM) model; |
| $M$ | → | A class of (ToM) models; |
| $\mathcal{M}$ | → | The (hypothesis) space[1] of all possible ToM models; |
| $f$ | → | A (ToM) process, mapping perceived environments $e$ to models $m$; |
| $o$ | → | A collection of insights about the environment (output(s) of a ToM model); |
| $\mathcal{O}$ | → | The hypothesis space of possible insights. |

Consider an Observer $\omega$ (a natural or artificial intelligence) observing an agent $a$ in a certain environment $e$.

**Definition 1.** *We term* Machine Theory of Mind *the problem of learning a mapping $f$ from the Cartesian product $\mathcal{A} \times \mathcal{E}$ (where $\mathcal{A}$ is the set of possible agents and $\mathcal{E}$ is the set of possible environment configurations), describing agents embedded in environments, to a space $\mathcal{M}$ of models*

$$f\colon \begin{array}{ccc} \mathcal{A} \times \mathcal{E} & \to & \mathcal{M} \\ (a, e) & \mapsto & m \end{array}$$

*where each model*

$$m\colon \begin{array}{ccc} \mathcal{E} & \to & \mathcal{O} \\ e & \mapsto & o \end{array}$$

*is itself a mapping from the perceived* environment $e$ *to a collection of* insights $o$ *about the agent $a$ present in $e$* [3] *(e.g., the intention of $a$, its most likely future decision, its emotional state, and so on).*

In an *open-world* formulation, the collection $\mathcal{E}$ of possible agent environments may also evolve over time, as the Observer expands its knowledge of the world, through exploration or teaching.

Following up from Principle 1, the mapping $f$ should be a function of time, as the Machine[4] Theory of Mind process (i.e., one's ability to build models of others' thinking) is continually (namely, episodically, at certaint time steps $\tau$, rather than continuously) updated in the light of new evidence (either time series of observations or rewards of interactions, $e(t)$, with $e$ and the agents therein)[5]:

$$f(\tau)\colon \begin{array}{ccc} \mathcal{A} \times \mathcal{E} & \to & \mathcal{M} \\ (a(t), e(t)) & \mapsto & m(t). \end{array}$$

An alternative, arguably more elegant, formulation separates the influence of the specifically observed agent and of its surrounding environment.

---

[3] Note that, in general, an environment *e* can contain more than one agent.
[4] In fact, this applies to both human and Machine ToM.
[5] Note that we use a different notation for the time instants τ in which the mapping is (asynchronously) updated.

**Definition 2**. *A Machine ToM learning process $f$ is defined as a mapping from (classes of) agents to models:*

$$f(\tau): \quad 2^{\mathcal{A}} \rightarrow \mathcal{M}$$
$$A(t) \mapsto m(t)$$

*(i.e., the shape of the model at any given time $t$ is a function of the agent's class only; this is not restrictive at all, as any class of agents in an hierarchical ladder can be considered, including individual agents/people the Observer $\omega$ knows very well), whereas*

$$m(t) = f(\tau)[a(t)]: \quad \mathcal{E} \rightarrow \mathcal{O}$$
$$e(t) \mapsto o(t)$$

*the current model $m(t)$, obtained by applying the current version of the Observer's ToM process $f(\tau)$ to the observed agent $a(t)$, takes as input the description of the environment (scene) $e(t)$ and maps it to a set $o(t)$ of insights about $a(t)$.*

The Machine Theory of Mind problem then becomes that of continually learning a process $f$ which, given an agent $a$ embedded in a dynamic context $\{e(t)\}$, encompassing a (long or short) history of environment configurations, generates a (ToM) model $m$ able to provide insights $o$ about that agent in that context.

## 3.2 Features of the ToM learning process

Our position is that the features of the Machine Theory of Mind learning process $f$ should mirror the Principles stated above, which, as we illustrated, are strongly backed up by the existing body of evidence in the literature:

1. It is a lifelong, continual learning experience (Principle 1).
2. This lifelong learning process is kick-started by a simulation of the self (Principle 2), at least if the other agent is deemed similar enough to us.
3. It consists on a sequential updating process which leverages both observations (in a supervised learning framework) and interactions with the environment, leading to rewards or penalties, in a reinforcement learning setting (or, more broadly) a sequential decision making one (Principle 3).
4. It should combine both a simulation of the other (following ST), and a mechanism for validating hypotheses about their reasoning process (in accordance with TT) (Principle 4).
5. It should incorporate an active learning (Principle 5) component, as part of the Observer's interaction with the environment.
6. It should be able to handle higher-order ToM (Principle 6).
7. It is a multi-task learning problem, as multiple insights about agents are sought (Principle 7), depending on the specific situation.
8. It should model the severe uncertainty associated with the intrinsic inaccessibility of another agent's mental states and feelings (Principle 8).

Incidentally, a number of research questions (RQs) are asked in[196] some of which are indeed related to the above Principles. For instance, RQ1 asks what theory underlies the use of the theory of mind (see our discussion of ST/TT/hybrid approaches). RQ2 asks which mental attitudes are modelled by the work (clearly about the choice of "insights" a ToM model should produce) while RQ3 wonders how does the system model knowledge related to the theory of mind (thus relating to the choice of cognitive models and simulation architectures. Interestingly, RQ4 asks whether a same computational model can describe both human and artificial agents. Still, to the best of our knowledge, no papers explicitly formulate requirements for Machine ToM such as those proposed here.

Next, we critical review the prior art to assess to what extent it adheres to these Principles.

# 4 Critical review of the state of the art

The main strands of latest approaches to Machine Theory of Mind have been briefly illustrated in Section 2. Here we screen them against the above requirements, at the level of whole frameworks, while reserving the right to scrutinised the most original or difficult to classify contributions individually.

A recent comparative analysis of theory of mind abilities in machine learning models can be found in[197].

## 4.1 Approaches vs principles

### 4.1.1 Principle 1: Lifelong learning

Among the main machine ToM strands, *(inverse) reinforcement learning*, as a sequential decision making setting, is quite naturally placed to address Principle 1. Outside this niche, however, papers by and large fail to address this important requirement. An exception, lifelong *cultural* learning using ToM is explicitly addressed in[198]. This is quite stunning, considering the importance of this element, and the vast and growing body of work on continual, online and lifelong learning within machine learning at large[199].

Consistently with this picture, few or no *Bayesian ToM* methods couple their focus on uncertainty (Principle 8) with a proper framework for lifelong learning (Principle 1): a partial exception, is, from a reinforcement learning stance, [200]. This is a clear gap which can and should be suitably addressed in the future, as continual Bayesian frameworks already exist outside of this particular area[201,202].

A lifespan perspective has been considered within the field of cognitive science[203,204,205,206]; yet, AI and machine learning are still to onboard any insights to the matter[207]. *Brain-inspired models* of ToM, e.g.[208], do consider learning (both supervised and unsupervised), but not quite how this should be a continuous process.

A very interesting paper is[209], where a number of biological mechanisms are described (both neuronal and non-neuronal) that may help explain how organisms tackle lifelong learning, and examples of biologically-inspired models and biologically-plausible mechanisms that have been applied to artificial systems are presented. Unfortunately, the discussion is not specific to artificial ToM.

*Self-improvement* is a rapidly growing area within Agentic AI[210,211], in which techniques such as Bayesian meta-learning[212], workflow optimisation[213], self-reflection[214], contextual adaptation[215] or evolutionary learning[216] are employed to continually adapt the agent to its environment and task. If employed for machine ToM, this clearly has the potential to systematically address the all-important Principle 1.

### 4.1.2 Principle 2: Leveraging own thinking

This principle is widely acknowledged in the cognitive literature[217,218,219]. However, it is unclear to what extent this has been brought on board by work on machine ToM. For instance, while Bayesian methods' potential for kickstarting the modelling from the self (Principle 2) is there[220], no actual papers seem to exist in this sense.

On the other hand, brain-inspired models do explicitly acknowledge learning from self-experience, e.g., via the Superior Temporal Sulcus (STS) [221]. The *Consequence Engine* proposed in[222] also provides a robot with the ability to self-model and hence predict the consequences of its own actions, in order to the predict another agent's actions. Self-explaining agents are also considered in[223]. Overall, this class of models is best placed to incorporare a simulation of the self into machine ToM; the integration of some form of biological/cognitive model is thus desirable, and one of the features of our proposed holistic meta-model.

As it deals with modelling the consequences of agent decisions in its environment, RL-based methods do not quite meet Principle 2, as they are entirely externally-focused.

As for LLMs, their advanced reflection capabilities can be employed for self-analysis, as it has been shown in several papers. One topic of study there is *self-alignment*, i.e., the ability of the model to align themselves without fine-tuning[224,225], which can be considered as a form of self-reflection. *Self-cognition*[226], defined as "an ability of LLMs to identify their identities as AI models, recognize their identity and demonstrate an understanding of themselves." is surveyed in[227]. Such an "understanding of themselves" is clearly the first step towards large language and foundation models employing self-simulation to kickstart a theory of other minds: this is an exciting research avenue for future work in this field.

### 4.1.3 Principle 3: Learning from observations and interactions

Principle 3 is at least partly met by reinforcement learning efforts[228,229], as there learning from environment interactions is central. Still, a coherent framework for learning from both observations[230] *and* interactions appears to be missing.

As for the other main strands, most Bayesian ToM methods rely on updating the model purely based on new observations (see, for instance, where an agent's degree of belief is updated as in $b_t(y) \propto P(o_t|x_t, y)b_t(y)$, where $P(o_t|x_t, y)$is the conditional probability of observation $o_t$ given the location $x_t$ and the state of the environment $y$. Observations are used to update beliefs in[231] as well. However, there are exceptions such as[232] (albeit outside the Theory of Mind domain), where beliefs are updated on the basis of actions taken by the Observer or by other agents. An interesting paper is[233], where a language-augmented Bayesian theory-of-mind (LaBToM) translates natural language into an epistemic ''language-of-thought'' with grammar-constrained LLM decoding; these translations are then evaluated against the inferences produced by inverting a generative model of rational action and perception. There, belief update is conditional on current state and prior belief, in a partially observable Markov decision process (POMDP) framework. All in all, Bayesian RL for Theory of Mind does have the potential to meet Principle 3.

Cognitive model-inspired methods and biology-inspired methods, on the other hand, as they attempt to mimick the cognitive or physical structure of the brain, do not concern themselves with the way the agent in its entirety interacts with its environment.

Finally, large language models can potentially learn from interactions when "scaffolded" in an agentic AI setting[234,235]. There, a foundation model (e.g., and LLM) acts as the core of an agent pursuing complex goals and planning, able to continually self-improve by interacting with the environment[236]. As mentioned, agent AI for ToM is a nascent field of investigation[237] which, however, has the potential to address both Principle 1 and Principle 3.

### 4.1.4 Principle 4: Marrying Simulation-Theory and Theory-Theory formulations

There is awareness in the community of the need to marry a Simulation and a Theory approach to ToM in order to achieve an effective MToM, or even move beyond this dichotomy. In[238], the authors claim TT has an advantage over ST in certain context (e.g., regarding agent development). The need for interpretability and the integration of neurosymbolic architectures is highlighted in[239]. Indeed, some authors associate TT with symbolic reasoning, and ST with pure learning from data, so that hybrid approaches should consider neurosymbolic reasoning as a suitable framework[240]. The meta-model we present in Sec. 5 is agnostic to the choice of the class of the (hybrid) model.

Almost all existing approaches, at any rate, adopt one or the other of the two stances.

A notable exception is[241], in which a novel approach is proposed that leverages Theory-Theory and Simulation-Theory to enhance ToM within the multi-agent reinforcement learning framework, making use of digital twins. Frameworks like those outlined by Zhinin-Vera et al. and the team at the Bristol Robotics Laboratory[6] do combine the two into hybrid systems. Finally, certain implementations of virtual agents use TT-based rule systems to govern social norms or states, alongside ST to mentally simulate the beliefs and goals of other actors, in a belief-desire-intention setting[242]. In[243], a comparison between the two frameworks is run, but without any attempt of integrating the two. Overall, a tiny majority of the research in this area appears to meet Principle 4.

[6] https://www.bristolroboticslab.com/

### 4.1.5 Principle 5: Active learning

When an AI tries to infer what a human wants simply by watching them (e.g., as in inverse RL), it faces a major bottleneck: many different internal desires could explain the same human behavior. To solve this, the AI may use active learning to intelligently prompt or query the human (e.g., "Would you prefer path A or path B?"). This targeted questioning can then allow the machine to rapidly map out and understand the human's mind. However, this potential seems to be currently severely undertapped.

While the wider role of humans in the loop in ToM is discussed in[244], and the role of higher-order cognitive models in active learning is touch upon in[245], there ToM is mentioned *en passant* but is not the focus of the work. Several application papers stress the importance of ToM for active learning in people, e.g., people with autism[246] or for teaching purposes[247][248], rather than active learning as a feature of a Machine Theory of Mind. Active learning and ToM are combined, in this sense, in[249] for marine teaching.

A notable exception within this landscape is[250], where active learning and continual learning are actually used within evolutionary computation to deliver computation theory of mind.

### 4.1.6 Principle 6: Higher-order ToM capabilities

In the inverse reinforcement learning approach to ToM, for instance, higher ToM levels[251] are modelled using *Interactive POMDPs*[252], in which other agents are modelled as part of the world model (the environment). Higher-order policies are considered within game-theoretical ToM[253,254], via the modelling of different value functions. In[255], a *colored trails* setting is used to determine the usefulness of higher-order ToM, where agents with iterated bestresponse beliefs (IBR) and utility-proportional beliefs (UPB) agents are compared.

Some attempts have been made to integrate recursive formulations into LLMs[256]. However, due to the black-box nature of LLMs, this remains a challenge. In[257], a combination of computational agents and Bayesian model selection is used to determine to what extent people make use of higher-order theory of mind reasoning in competitive games. A benchmark for testing higher-order also ToM exists, named Hi-ToM[258] (see Sec. 6).

### 4.1.7 Principle 7: Multi-task nature

It is fair to say that most, if not all contributions to machine ToM recognise the fact that Theory of Mind is an umbrella term subsuming a variety of different tasks, e.g., estimating or predicting desires, beliefs, intentions, false beliefs, emotions, percept. As such, Principle 6 and the need to provide a variety of "insights" on the other agent is well recognised. However, it is also fair to say that most if not all focus (quite understandably) on only a subset of those. Research typically uses four to five primary cognitive categories, which are assessed using testing batteries containing anywhere from 4 to 40+ specific scenarios and subtasks[259]. More attention, however, needs to be paid to the *representations* of those insights, as highlighted in our formal definition (Sec. 3).

Interesting recent papers *explicitly* acknowledging the need for a multi-task approach is, for instance, based on a distributed multi-task meta ME-AIRL framework based on theory of mind and mean field[260], where *mean-field theory* is key to transform interactions between complex tasks into interactions between the main task and the average of the remaining tasks. Within robotics, Yu et al.[261] have designed a dynamic trust-aware reward function based on different trust beliefs to guide the robot policy learning.

### 4.1.8 Principle 8: Acknowledging uncertainty

Bayesian (machine) ToM approaches [262,263] rely on a probabilistic representation of agents' mental states, including beliefs and desires, e.g., by conceptualising agents' planning and inference processes[264] as partially observable

Markov decision processes[265]. As such, they clearly meet Principle 7, as they have the potential to represent the epistemic uncertainty involved.

Alternative ways of quantifying uncertainty in ToM have also been proposed, e.g., in[266], where frequentist statistics is employed, or in multi-agent systems in[267], where a probabilistic (but not Bayesian) representation is used. Uncertain beliefs are also considered in[268], in which language models are asked to formulate how uncertain they are abour their predictions, using methods such as *calibration* or *direct forecasting*[269].

Importantly, *second-order uncertainty*, which has proven itself quite effective in recent times within machine learning at large[270,271,272], using either random set[273] or credal set[274,275,276] representations, has never been employed for theory of mind modelling to date, leaving an important opportunity to exploit for researchers in the near future.

# 5 A holistic meta-model for machine theory of mind

To address the above limitations, here we propose a *meta-model* for Machine Theory of Mind, as a blueprint for the development of future computational models adherent to our stated, evidence-backed principles.

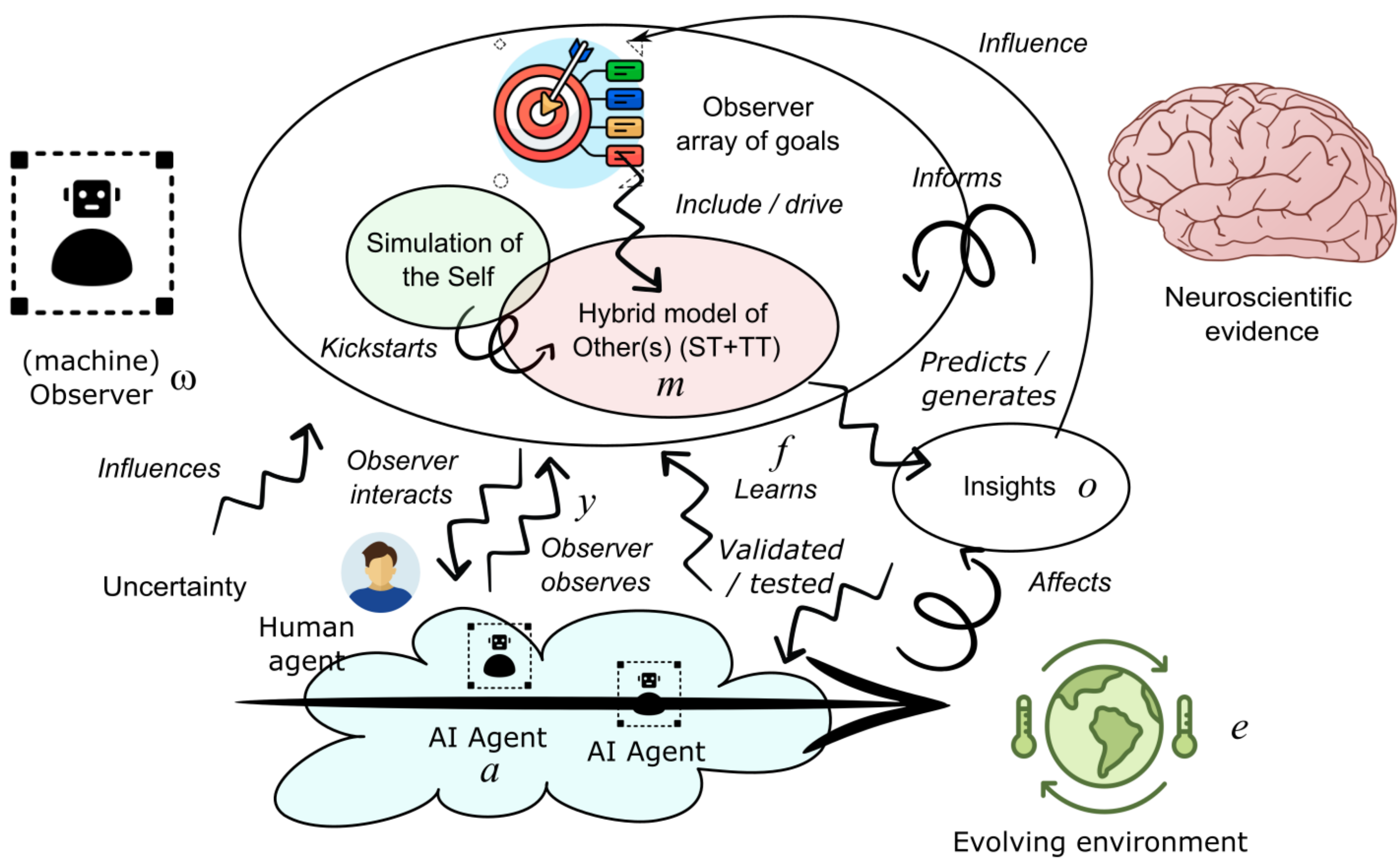

**Figure 1**. *Illustration of the proposed meta-model for a Machine Theory of Mind. A Hybrid Model of the other Agents in an evolving Environment, influenced by the available neuroscientific and cognitive knowledge, is kickstarted by the simulation of the self. In a dynamic process in which the ego agent (Observer) interacts with the evolving Environment, the Hybrid Model produces insights on the other agents (artificial or human) present there, which the Observer observes but also interacts with. Feedback on the quality of the predictions / insights drives the evolution of the core model for the relevant agents of classes thereof, but also the generation of new hypotheses on the others' behaviour to be tested. ToM objectives are only a subset of the Observer's whole set of goals. Uncertainty affects of all elements of the picture: perception, simulation, hypothesi testing, nature and quality of the insights*.

## 5.1 Overall model

The proposed meta-model is a model of the ToM-endowed Observer *as a whole*, as there is evidence that ToM-related goals and actions are closely intertwined with any other goals the Observer might have[277,278,279,280,281].

In agreement with our principles, the Observer both *observes* and *acts in* a dynamic, evolving environment, which contains other intelligent agents (artificial or human, see Figure 1). Its actions and observations are, in general, determined by the Observer's wider goals (not necessarily ToM-related[282]) – they may at times appear to happen for no specific reason.

Indeed, having better interactions with the environment (namely, with the agents therein) is the very reason for existence of theory of mind, both in humans[283] (with implications to the evolution of language[284]) and in machines[285]. The Observer (Figure 1) has a set of own goals (some of which, but not all, are related to Theory of Mind), which in turn determine the way it interacts with the environment (in fact, this applies, once again, to both humans and machines).

Our proposed meta-model rests on a Hybrid model of the other(s) compounding, in its most general formulation, both Simulation-Theory and a Theory-Theory stance to theory of mind (Principle 4).[7]

### 5.1.1 Simulation aspect

The core Theory of Mind component of the Observer needs to include a process for dynamically learning and deploying (Principle 1) a simulation of the other agent(s) in the environment. The ToM process (see Section 3) consists of a mapping from the environment configuration to a learnable model of some kind. We kept our stance fully general there, to avoid unnecessarily restricting the possible range of model classes.

Principle 4, however, provides evidence in support of the model $m$ to assume the form of a *simulation architecture*, as the simulation-theory component of the overall meta-model. Its features are discussed in more detail in Section 5.2. Nevertheless, as explained, our proposed general formulation is broader than this.

Importantly, the learning of the most accurate simulation for a given agent is initialised by the Observer's simulation of itself (Principle 2) – over time, the accumulation of evidence leads to a bespoke simulation for classes of agents. The usefulness of "stereotypes" in social cognition for quickly understanding "types" of people we do not know well and have no time to know better is supported by the literature[286,287]. Formally, this can be represented as the learning of a ToM model for a *class* $A$ of agents, rather than an individual agent $a$, which can be very efficient in contexts such as, e.g., autonomous driving in which quick decisions need to be made. As the amount of evidence about individual agents grows in the MToM continual learning process, individuals (singleton classes $A = \{a\}$) with which it has frequent interactions can start being modelled. One can argue that the simulation is potentially always there (dormant), even when the Observer has no interaction with that particular agent. An updating of the simulation only takes place if and when the agent of interest is detected in the environment[288,289,290].

According to the principle of cognitive economics[291], we want (for efficiency reasons, rather than adherence to the highlighted Principles) tailored simulations for individuals or groups to be assembled from the same underlying building blocks (as neuroscientific evidence seems to suggest is the case for the human brain). This ties (as we will see below) with the thorny issue of complexity[292] which chatacterises general (evolutionary[293] or progressive[294]) neural architecture search (NAS)[295] approaches to neural (simulation) learning (although polynomial complexity[296] and other efficient approaches[297] have been formulated).

Based on Principle 3, the simulation is updated based on both Observer interactions and observations, and provides insights (Principle 7) about each of the agents in the environment.

[7] In fact, it is flexible enough to be extended to approaches outside this dichotomy, e.g., perception theory.

### 5.1.2 Theory aspect

Following from Principle 5, the meta-model must also contain a theory-theory (TT)-inspired component. As mentioned in Sec. 1.1, some authors interpret TT as basically the existence of an innate brain cognitive model: in this sense, the injection of a-priori knowledge related to the most suitable model/simulation architecture can be seen as a form of TT. Gopnik and Wellman[8], instead, use a popular "scientist" metaphore in which theories should: (i) involve and appeal to abstract unobserved entities; (ii) invoke explanations in terms of these abstract entities and laws; (iii) lead to predictions, and (iv) lead to distinctive interpretations of the evidence. The shape these theories take, or what kind of hypotheses testing should be applied, remains vague, leaving flexibility to design a holistic metamodel like ours in which the hybrid ToM model features aspects of TT. As mentioned in Sec. 4.1.4, in[298] it is suggested that TT is related to symbolic reasoning (a knowledge base that consists of beliefs of others' mental attitudes), ST to learning, and hybrid approaches to neurosymbolic AI – the latter also supported in[299]. So, a neurosymbolic approach to the hybrid model component of our metamodel should be encouraged.

TT can also be incorporated as a form of "dynamic tuning" of the machine's own goals, which are also assumed to evolve over time (in line with recent online learning work[300]). This, as discussed in Sec. 2, is driven by the uncertainty attached to the available insights on the other agent(s) (Principle 8). Indeed, in the presence of ambiguous insights (ToM outputs), the Observer may want to focus its observations or put actions in place to reduce such ambiguity and learn, in the process, how to improve its own ToM abilities. The notion of evolving Observer goals is further elaborated upon in Sec. 5.2.3.

**Generation of ToM goals for the Observer**

Rather than by explicitly formulating hypotheses to be validated, as in a classical TT approach (e.g., is my friend tapping on the table because he has work troubles, or only because he enjoys the music?), one option is to add a suitable component to the Observer's set of objective / reward functions (in a multi-objective reinforcement learning setting[301]), driving it to maximise several, possibly conflicting and/or incommensurable objectives.

A number of possible ToM-related objectives (associated with reward components) can be imagined, e.g., achieving a "better" interaction with the other[302] (measured by some "quality" index[303]), acting to lower the (aleatoric[304] or epistemic[305]) uncertainty about the generated insights, or achieving a superior-quality simulation of the other agent(s) (measured by more accurate predictions of their future behaviour, which is observable). In this set-up, active learning is a by-product of the feedback from the insights on the environment to the set of Observer goals ("affect" link).

**Generation of ToM hypotheses**

The other option is for the Hybrid model (the core of our holistic Observer representation) to actually formulate hypotheses to be tested: as illustrated in the literature, these hypotheses do fall in the category of what we called ToM "insights". How these can contribute to the ToM learning process is discussed in Sec. 5.2.3.

Finally, a possible, very exciting approach is to adopt (epistemic[306]) diffusion models, generative adversarial networks (GAN)[307] or variational auto-encoders (VAE)[308] (or any other suitable generative architecture) to generate data and/or hypotheses to disambiguate uncertainty in a more classical TT stance: this is something that had not been attempted so far and would be important to explore in future work.

---

[8] Note that the dynamic feature of theories (see Sec. 4.1.1 and 5.2.3 below) is already acknowledged there, page 262.

## 5.2 Components

The core of the model is the ability, built over time, to simulate the mental life of other agents (Principle 2, pink bubble in Figure 1), kickstarted by a simulation of the self (Principle 1, green bubble). This simulation is affected by the uncertainty that comes from an imprecise or incomplete knowledge of the dynamic environment (Principle 8, light blu cloud at the bottom).

Delving deeper, there are four main components (not necessarily in this order) to a our holistic metamodel to a computational theory of mind:

1. The **simulation approach** (e.g., NAS, *neural architecture search*, versus NMS, *neural module search*[309,310,311:] see Sec. 5.2.1).
2. The **search space** for the ToM simulation architecture (the class of potential structures of the sought simulation, Sec. 5.2.2).
3. The **learning process** by which the holistic model (and, in particular, its internal simulation architecture) is continually updated based on the available data (Sec. 5.2.3).
4. The **modelling of the uncertainty** induced by latent mental states, unpredictability, and scarcity of data (Section 5.2.4).

Under the simulation stance we have two options: (1) one simulation per scene, e.g., one simulation for the ensemble of multiple agents presents in the scene; (2) one simulation for each individual agent in the scene. The very nature of the concept of theory of mind, however, suggests a simulation should be specific to each agent, as it related to its own specific mental world[312], in accordance with both Definition 1 and Definition 2.

### 5.2.1 Simulation strategies

Several ways of learning the best neural architecture for solving a given task have been proposed within machine learning (ML), which have the potential to be employed for kickstarting and updating a ToM simulation of other agents. We briefly recall them here.

*Hyperparameter optimisation* is a central element of ML, as the problem of choosing the optimal values of a hyperparameter (i.e., a parameter controlling the learning process) for a learning algorithm. Approaches such as grid search, random search, Bayesian, gradient-based and evolutionary optimisation have all been explored.

*Neural architecture search (NAS)* is a technique for automating the design of artificial neural networks (ANN), a widely used model in the field of machine learning, and goes beyond hyperparameter optimisation[313] to discover entirely new architectures. NAS has been used to design networks that are on par with or outperform hand-designed architectures[314,315]. Methods for NAS can be categorised according to the search space, search strategy and performance estimation strategy used[9]:

- The *search space* defines the type(s) of ANN that can be designed and optimized.
- The *search strategy* defines the approach used to explore the search space.
- The *performance estimation strategy* evaluates the performance of a possible ANN from its design (without constructing and training it).

*Continual neural architecture search* attempts to bring together NAS and continual learning[316], including class-incremental learning[317], in an effort to allow a flexible evolution of neural structures which need to adapt to evolving tasks and environments. A relative unexplored field of study composed by an assemblage of different methods mutuated from traditional continual learning[318], its formal definition is still uncertain, but it aims to be domain agnostic, dynamic while having bounded capacity and learning rule plasticity.

[9] https://en.wikipedia.org/wiki/Neural_architecture_search

*Neural module search* is a more specific approach, in which a natural language parser is used to dynamically lay out a deep network composed of reusable modules for visual question answering. Despite the original domain application, it seems applicable to a reconfigurable ToM simulation approach in which basic neural modules are flexibly (re-)connected to generate simulations of different (classes of) agents. Another example is[319], which attempts to overcome the computation and time costs of classical NAS by (very relevantly to the topic of this paper) injecting knowledge into the search. The authors do so by proposing a ModuleNet model which inherits knowledge from existing convolutional neural networks (CNN) decomposed into modules rearranged into a knowledge base. One can regard NMS as just a special case of NAS in which some parameters (corresponding to the base modules) are frozen, which can potentially address the dreadful computational complexity of full-blown NAS, which is only complicated by the online learning needs of machine theory of mind (see Sec. 5.2.3). Modular approaches are also supported[320].

*Meta-learning*[321] is another terms of relatively vague defitinition, but as whole it relates to the notion of using any available metadata to understand how automatic learning can become flexible in solving learning problems. An alternative name for it is *learning to learn*[322]. Three requirements need to be satisfied: (1) the system must include a learning subsystem; (2) experience is gained by exploiting meta knowledge extracted in previous learning episodes, possible from different domains; (3) the learning or *inductive bias*[323] (i.e., the set of assumptions made by the designer of the model about the data) must be chosen dynamically.

*Auto-ML*[324,325] is possibly the most general umbrella terms, as it aims to fully automate the tasks of applying machine learning to real-world applications. As such, it can be considered to include all of the above sub-areas.

*Other approaches* include online convex optimisation[326] (a sequential decision making framework in which an online player iteratively makes decisions, whose associated outcomes are unknown to the player and cause a loss to the decision maker, also unknown to the decision maker beforehand) and online learning[327]. A nice illustration of the comparison between adaptive learning and the various machine learning paradigms described above can be found in[328], Figure 1.

Overall, continual neural architecture search (possibly made less expensive and general by imposing suitable constraints, as in neural module search) appears to be the best suited mechanism for learning and updating a ToM simulation. The question then becomes how to properly design a search space for a (reconfigurable) neural ToM simulation.

### 5.2.2 Search space for simulation architectures

As shown in Figure 1, *neuroscientific evidence* can be very useful to help us design the search space of simulation architectures[329]: this, implicitly, is the philosophy underlying biologically-inspired models[330,331,332,333], including spiking neural networks[334,335].

Within this framing of the problem, the principle of cognitive economics drives a push for tailored simulations for individuals or groups to be assembled from the same underlying building blocks, supporting *module-based architectures* which allows us to reduce the parameter space of the search. There is indeed substantial neurological and developmental evidence of assemblage that the human brain achieves Theory of Mind by assembling basic, foundational building blocks, rather than possessing it as a single, innate module[336,337,338]. Further evidence of the assemblage of basic building blocks from brain studies exist outside the context of theory of mind[339,340,341], reinforcing this additional principle.

Neuroscientific evidence, however, is not the only source of constraints for a suitable search space of neural ToM simulations. Hierarchies of mental states have been proposed in cognitive psychology[342,343], for instance in a Bayesian inference setting similar to graphical models[344], often in other contexts like delusion[345]. Definition 1, however, with its multiple outcomes of interest ("insights"), resounds reasoning on graphs[346], in which any (quantity associated with any) node in the graph structure can be the target of the inference. This might indicate that *a graph structure, rather than a hierarchy*, could be more suitable or more general in scope[347]. As in GNNs, multiple tasks associated with different nodes[348,349] could be the objective.

Based on this, one can envisage various possible choices for the class of simulation architectures: hierarchies of mental states in a Bayesian framework; (Bayesian, or event epistemic[350]) graphs of mental states; inspired by mPFC / TPJ descriptions, as in[351,352]; *belief-desire-intention* (BDI) models[353], which have been widely explored in machine theory of mind in the past[354], plugged in, however, into our proposed comprehensive meta-model setting.

Our proposed meta-model is completely agnostic to the source of knowledge informing the search space of the simulation, leaving the door open to future developments in the sister disciplines.

### 5.2.3 Learning framework: multi-objective agent goal optimisation

In our proposed meta-model, the learning aspect of ToM is not isolated from the learning process the whole Observer agent $\omega$ goes through. Rather, various types of ToM goals constitute part of the agent's overall set of goals, which the Observer attempts to optimise as a whole, in *multi-objective optimisation*[355,356] framework. There is no arguing with the point that agents possess multiple, often incommensurable objectives[357], including Theory of Mind – so, this is a very general formulation of the holistic agent learning problem.

**The holistic optimisation problem**

Formally, the latter can be posed as:

$$arg \max_{\theta \in \Theta} h(g_1(t), \dots, g_{N(t)}(t)),$$

where $g_1, \dots, g_{N(t)}(t)$ are $N(t)$ (time-varying) objective functions, possibly conflicting and/or incommensurable, the agent strives to optimise (at a given time $t$), $\theta \in \Theta$ are the model parameters in a certain hypothesis space $\Theta$, and $h$ is an arbitrary function which compounds all partial objectives. To be as general as possible, $h$ can assume any form: in multi-objective optimisation, common choices are a weighted linear combination of the partial objectives[358], or an exponential function of the latter[359].

This needs to be thought of as a continual learning process, as stated in Principle 1.

Some of those objective functions are related to Theory of Mind; in particular, one can formulate four broad classes of such ToM objectives[10]:

- A *supervised* objective, which measures how accurate are the *predictions* (observable insights) made by the ToM model: $g_{ToM-SL} = \int_{t=1,\dots,T} \| o_i - y_i \|$, where $o_i$ is an observable insight, $y_i$ is the corresponding measured *observation* (within the environment $e$), $\| \ \|$ is an arbitrary norm in the space of observations, and the integral cumulates discrepancies over a certain period of time $t = 1, \dots, T$.
  Minimising $g_{ToM-SL}$ produces models which generate optimal predictions; using the this signal as feedback for training results in continually updating the model (whatever hybrid of ST and TT the latter is).

[10] Although others are possible, and this does invalidate the generality of our meta-model.

- A *reinfocement learning* objective, which measures how effective the decision-making strategy of the Observer is, when it comes to maximising a suitable *cumulative expected reward* over a certain period of time. Within RL, this is normally denoted as

$$g_{ToM-RL} = G_t = \sum_{k=1}^{\infty} \gamma_k R_{t+k+1}$$

  where $R_{t+k+1}$ is the future reward received at time $t+k+1$ and $\gamma_k$ is a *discount factor* which attenuates the weight of future decisions. For ToM purposes, a number of possible shapes for the reward function can be envisioned (and have actually been implemented in the literature): effective collaboration between the Observer and the other agent(s), quality of their interaction, and so on.
  Maximising $g_{ToM-RL}$ delivers an optimal series of decisions/actions for the Observer (optimal in the sense determined by the chosen reward); using it for training produces the parameters of the *Q function*[360] (if a model-free approach is followed) or of the underlying *Markov Decision Process*[361] (in model-based RL).
- A *hypothesis generation* objective, which measures whether the hypotheses generated by the model are true to a given statistical confidence level: $g_{ToM-TT} = 1(hypothesis\ is\ true)$. This can be done, for instance, by setting the problem as a statistical hypothesis testing[362] one, in which one needs to decide if a data sample is typical or atypical compared to a population assuming a hypothesis we formulated about the population is true. Epistemic extensions of statistical hypothesis testing using random sets, rather than probability distributions, are also possible[363] (see Sec. 5.2.4).
  Maximising $g_{ToM-TT}$ delivers a model that generates statistically correct hypotheses.
- An uncertainty reduction objective, which strives to minimise the uncertainty attached to a given insight $o_i$, i.e.: $g_{ToM-U} = \int_{t=1,\dots,T} u(o_i)$, where $u(.)$ is a measure of (aleatoric, epistemic or total) uncertainty[364].
  Updating the model to minimise $g_{ToM-U}$ leads to a model whose insights are, over time, less and less uncertain and therefore more reliable.

Practical questions that need to be addressed by any computational implementations of our meta-model are, amongst others: what relation is there between the continual and reinforcement learning components of the model (as RL already a form of continual learning[365,366,11]), and what is the precise nature of the relevant signals (e.g., observations, actions, active probing, and so on).

**Dynamic nature of Observer goals**

It is important to point out that, in the most general setting, *the set of Observer goals itself, $g_1, \dots, g_{N(t)}(t)$, is bound to evolve with time.* For instance, $g_{ToM-TT}$ will vary with the hypothesis to be tested; $g_{ToM-U}$ will vary with the particular insight affected by the excessive level of uncertainty the Observer wishes to reduce.

This Evolving AI concept is illustrated in Figure 2. Exploring the interrelation between it and (continual) Machine Theory of Mind is an exciting topic for future research.

This is in line with a very recent push (including by us) towards an *Evolving Machine Intelligence* paradigm (only partially address by paradigms such as online learning[367] or online convex optimisation[368]), in which the AI's (in our case, the Observer's) meta-objective to learn from, interact with and understand the environment drives the emergence of new goals, in response to ever-changing requirements as well as the drive to improve its current understanding of the world's (causal) structure. New goals can emerge by continually exploring the space of possible objectives, encoded at both a "semantic" level (e.g., the AI realises it needs to gather more data about a certaint agent) and a "functional space" level, the concrete mathematical incarnation of semantic goals (e.g., what objective function/constraints do I optimise to achieve that?). A compositional strategy can then be sought to decompose objectives into steps to make the problem tractable.

[11] https://www.youtube.com/watch?v=JSatkDPx2w4

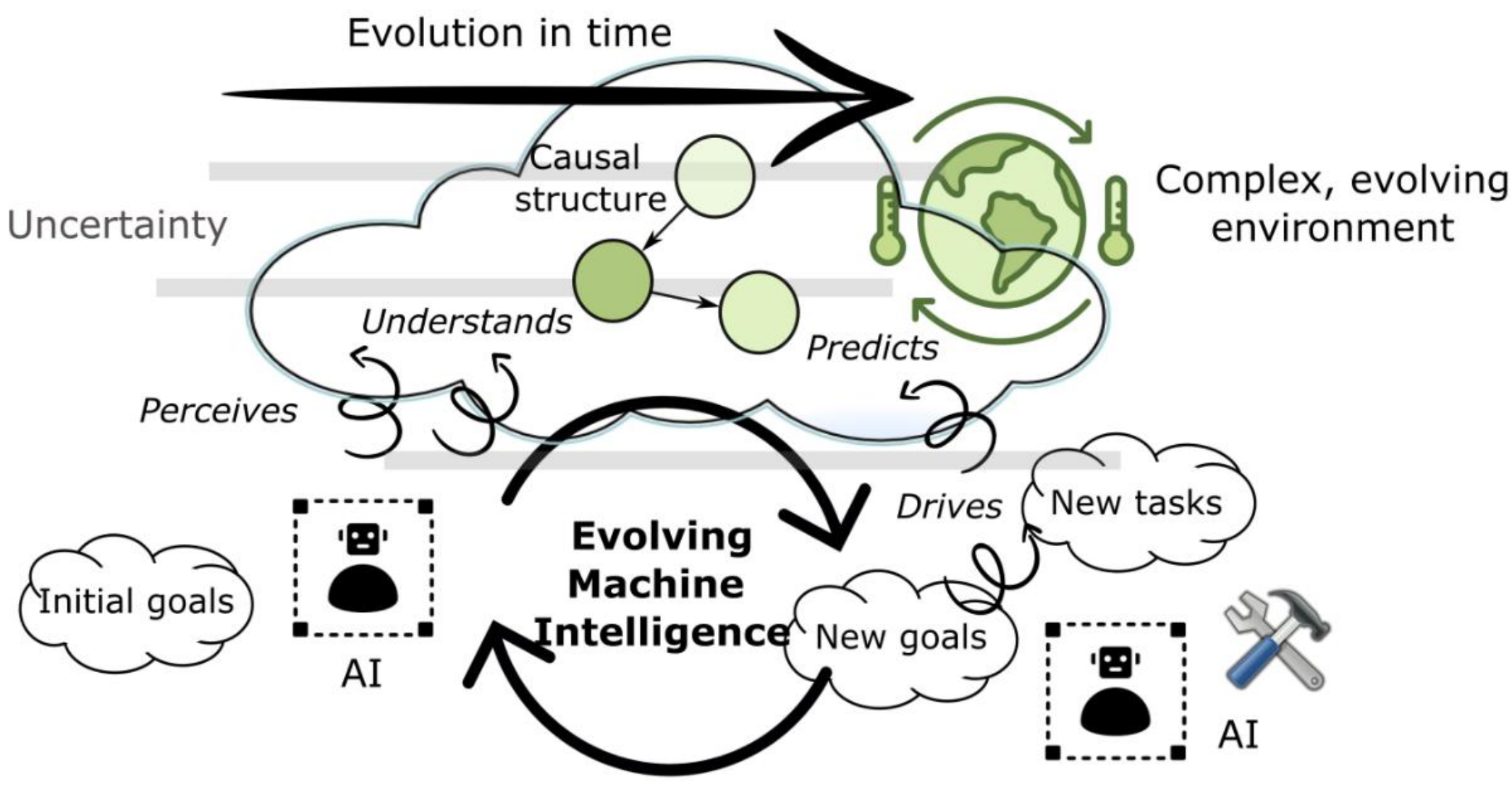

**Figure 2.** *Concept of Evolving Artificial Intelligence*.

### 5.2.4 Modelling of uncertainty

**Methods**

Epistemic uncertainty should be integral to the model, because of the extreme uncertainty of guessing another person's mind from limited cues (Principle 8). Within the wider deep learning and machine learning community, the most popular approaches to uncertainty quantification are currently Bayesian deep learning[369], ensemble learning[370], conformal learning[371], interval neural networks[372] and evidential neural networks (INNs)[373], together with the rising field of second-order, epistemic deep learning. An excellent recent review on the subject can be found in[374].

In principle, all of those are applicable to quantifying uncertainty in Machine Theory of Mind: however, it has been pointed out that, for instance, Bayesian and ensemble methods suffer from high computational complexity[375], while INNs can only supply deterministic intervals, rather than probabilities or higher-order uncertainty measures. Evidential approaches predicting Dirichlet second-order distributions appear to violate asymptotic conditions and struggle with out-of-distribution data[376]. Bayesian reasoning has been criticised because of need to (often arbitrarily) pick a prior distribution[377], and the difficulty of predicting a full Bayesian distribution (forcing designers to employ Bayesian Model Averaging[378] methods).

Second-order or "epistemic"[379,380] approaches emplying random-set[381] or credal set[382, 383,384,385]representations, on the other hand, have shown to outperform competitors in recent years in terms of accuracy, robustness, calibration, out-of-distribution detection, and to be able to provide tighter generalisation bounds[386], thus showing much promise when addressing the inherent uncertainty associated with latent mental states and unpredictable human behaviour.

**Components of the meta-model**

Which components of the meta-model are affected by uncertainty? Its effect is, in fact, present quite across the board.

*Observations.* First of all, the way the model observes the world is bound to be imperfect, whatever the sensory channels used. Text can be highly ambiguous (as attested by LLM experiments, see Sec. 2 and 6). Visual inputs are also know to be tricky, because of the intrinsic hardeness of solving even simple tasks such as identifying and detecting an intelligent agent within a visual scene: this can be due to a variety of nuisance factors (occlusions, illumination conditions, viewpoint, and so on).

The *highly dynamic nature of the Environment* is also a source of uncertainty, both from a cognitive[387] and an AI (e.g., reinforcement learning[388]) point of view. This is only starting to emerge as a recognised problem with continual machine learning, although awareness is on the rise[389].

*Human behaviour* is incredibly rich, due to our varied mental lives and the manifold ways our mental state can manifest themselves exteriorily[390,391], as also attested within neuroscience[392]. This produces a challenge for all the approaches listed above, including second-order "epistemic" probability methods, to be able to learn a wide enough range of alternatives for the detected for future human behaviour.

The latter point is reflected in the *structure of the ToM simulation* itself. Whether a (Bayesian) hierarchical model is used[393], with its conditional updating relations among levels, a BDI model[394], with its (probabilistic) logic implications[395,396] (e.g., in a ProbLog[397] style) between mental states, a graph[398,399,400] (where, again, reasoning take place via a message passing algorithm heavily influenced by epistemic uncertainty[401,402]) or any other (possibly module) structure discovered by (continual) neural architecture search[403] (as in FNAS[404]), both aleatoric and epistemic uncertainty are ubiquitous and condition almost all elements of the neural structure.

The *generation of new hypotheses*[405] and/or agent goals is also subject to uncertainty, in particular if formulated in natural language, as shown by recent attempts to extend transformer architectures and large language models to the credal[406] or the random-set[407] representation in order to improve their creativity and flexibility. This is a nascent problem in generative AI[408,409], but one that is likely to be very impactful in the future.

*Nature of the insights.* Last but not least, whatever our choice of representation for the multiple insights output by the machine ToM model, these are also bound to come an attached level of uncertainty. This is true for the classical probability distributions of classes generated by a classifier network[410,411] (e.g., a mental or emotional state or future intention) as well as for variables output by a regression model[412413] (e.g., a future trajectory) and even tensors[414].

## 5.3 Generality of the meta-model

We already proved above that our meta-model meets our eight evidence-backed Principles for a Machine Theory of Mind. It is worth stressing its generality as well, i.e., its ability to subsume all existing research strands as special cases.

While IRL approaches are based on a limited interpretation of Theory of Mind (i.e., that it consists of recovering the other agents' reward functions), the sequential decision making framework typical of RL is very much embedded in our model's continual interaction setting.
Game-theoretical theory of mind[415], as it rests on a similar formulation based on policies and higher-order value functions (or on quantal response equilibria[416]), also falls within the remit of our meta-model, in which the nature of the core ToM hybrid model is left as general as possible.
Bio-inspired computational models naturally adhere to a model in which expert knowledge from brain science is injected into the machinery simulating another agent's thinking. The nature of this injection (i.e., whether in the form of neurosymbolic reasoning, a bias on the choice of the architecture, or anything else) is not specified in our holitic model. The same holds for cognitive-inspired approaches (e.g., BDI), where prior knowledge is about abstract mental states, rather than brain regions which suitable functionalities.
Bayesian ToM models (whether in conjunction with POMDPs[417], BDI models[418,419], Bayesian "reciprocators" based on utility theory[420], or networks of Bayesian agents[421]) are also subsumed by a holistic model in which the modelling of uncertainty is central. Note that several efforts combine more than one of these strands into original approaches.

Finally, LLMs' black box nature makes it difficult to analyse their (emergent?[422,423,424]) internal structures, other than acknowledging their hierarchical transformer structure. Still, as our core "model" component is non-prescriptive, it is still general enough to encompass LLMs. The latter obviously do not "interact" with an environment if not verbally, unless scaffolded within a broader agenti architecture, which shows similarities with our holistic setting.

# 6 Benchmarking

Once having established the first formal definition of a Machine Theory of Mind, backed by principled evidenced in the four intertwining fields of cognitive psychology, neuroscience, artificial intelligence and machine learning, screened the existing state of the art efforts against these principles and definition, proposed an overall holistic meta-model for machine ToM, and discussed the possible design options to translate the meta-model into actual approaches, our excursus would not be complete without understanding how suitable existing machine ToM benchmarks are to address those challenges.

There is a vast literature on (mostly linguistic) tests within cognitive psychology (e.g.,[425], in particular using gamified environments[426]): this is not the place to recall that – rather, we focus on AI/machine learning benchmarks[427] (e.g., VQA for visual question answering[428]) which can be used to assess the viability of our metamodel, or a specific model derived from it. This an objective shared, for instance, by[429]. Another interesting recent paper[430] asks the question of whether Theory of Mind benchmarks need explicit human-like reasoning in language models.  Its results suggest that current ToM benchmarks may be solvable without requiring the explicit, human-like simulation of mental states they were designed to probe (and therefore unsuitable for true machine theory of mind investigations).

## 6.1 Recent efforts

We review the most recent (but also most established) benchmarks / datasets according to the nature of the data.

### 6.1.1 Language

An interesting recent benchmark is **FANToM**[431], which draws upon important theoretical requisites from psychology and necessary empirical considerations when evaluating large language models. Multiple types of questions are there formulated to challenge false sense of ToM capabilities in LLMs. The authors show that LLMs perform significantly worse than humans on FANToM. A very recent position paper goes further, to maintain that existing ToM datasets are simply "broken" for large language models[432].

Another recent banchmark is **OpenToM**[433], which uses longer narrative stories, characters with explicit personality traits, actions that are triggered by character intentions and questions designed to challenge LLMs' capabilities.

**ToMChallenges**[434] is based on Sally-Anne and Smarties tests with a diverse set of tasks, and includes an auto-grader to streamline the answer evaluation process. Three models were tested: davinci, turbo, and GPT-4. In related work, a plug-and-play multi-character belief tracker for LLMs is proposed in[435].

**ToMBench**[436] has three main features: a systematic evaluation framework encompassing 8 tasks and 31 abilities in social cognition, a multiple-choice question format to support automated and unbiased evaluation, and a build-from-scratch bilingual inventory to strictly avoid data leakage.

**Multi-ToM**[437] is an effort to evaluate multilingual theory of mind capabilities in large language models, by translating existing (textual) ToM datasets into multiple languages, effectively creating a multilingual ToM dataset and enriching these translations with culturally specific elements to reflect the social and cognitive scenarios relevant to diverse populations.

**EPITOME**[438] is a battery of six experiments that tap diverse ToM capacities, including belief attribution, emotional inference, and pragmatic reasoning.

**SUITE**[439] (Scaling Up Individualized Theory-of-Mind Evaluation) is a framework that enables large-scale, individualised ToM assessment of large language models using naturally occurring social discourse.

Quite uniquely, Higher-order ToM in LLMs, which involves recursive reasoning on others' beliefs, is the purpose of **Hi-ToM**, which also incorporates a new deception mechanism in ToM reasoning.

**RecToM**[440] is a novel benchmark for evaluating ToM abilities in recommendation dialogues. RecToM focuses on two complementary dimensions: cognitive inference and behavioral prediction. The former focus on understanding what has been communicated by inferring the underlying mental states. The latter emphasises what should be done next, evaluating whether LLMs can leverage these inferred mental states to predict, select, and assess appropriate dialogue strategies.

**AnaToM**[441] is a rather interesting and valuable new dataset generation *framework* for ToM evaluation named AnaToM. To realize an "Anatomy of Difficulty" in ToM reasoning, AnaToM strictly controls structural parameters such as the number of entities and the timeline in a story. This parameter control enables the isolation and identification of factors affecting the ToM of LLMs, allowing for a more precise examination of their reasoning mechanisms.

**UniToMBench**[442] is a recently introduced unified benchmark that integrates the strengths of SimToM and TOMBENCH to systematically improve and assess ToM capabilities in LLMs by integrating multi-interaction task designs and evolving story scenarios. Supported by a custom dataset of over 1,000 hand-written scenarios, UniToMBench combines perspective-taking techniques with diverse evaluation metrics to better stimulate social cognition in LLMs.

**PersuasiveToM**[443] is benchmark for evaluating machine theory of mind in persuasive dialogues, which considered two tasks: ToM Reasoning, which tests tracking of evolving desires, beliefs, and intentions; and ToM Application, which assesses the use of inferred mental states to predict and evaluate persuasion strategies.

**DialToM**[444] is a Theory of Mind benchmark for forecasting state-driven dialogue trajectories, composed on data from three domains of conversational data: *motivational interviewing* (AnnoMI[445]) sessions, *emotional support* conversations (ESConv[446]), and *persuasive conversations* (Persuasion For Good[447]).

**DebaToM**[448] is an innovative debate framework for Theory of Mind reasoning using Large Language Models. The DebaToM framework consists of a generator and a discriminator based on LLMs, which continuously optimise the reasoning process (see Principle 1) through multiple rounds of interactions.

**CogToM**[449] is a comprehensive, theoretically grounded benchmark comprising over 8,000 bilingual instances across 46 paradigms, validated by 49 human annotators. A systematic evaluation of 22 representative models, including frontier models like GPT-5.1 and Qwen3-Max, reveals significant performance heterogeneities and highlights persistent bottlenecks in specific dimensions.

**TOMATO**[450] consists of 5.4k questions, 753 conversations, and 15 personality trait patterns, and is used to evaluate nine LLMs, finding that even GPT-4o mini lags behind human performance, especially in understanding false beliefs.

**ToMI**[451] builds on the pre-existing ToM-bAbi dataset to generate stories with considerably more randomness, distractor phrases, distractor locations, distractor characters and a shuffling of the order of actions.

**BigToM**[452] is a social reasoning benchmark for LLMs which consists of 25 controls and 5,000 model-written evaluations.

**SimpleToM**[453] probes multiple levels of ToM reasoning, from mental state inference (explicit ToM) to behavior prediction and judgment (applied ToM). Second, it situates these tasks in diverse, everyday scenarios — such as supermarkets, hospitals, schools, and offices — where information asymmetries naturally arise (e.g., hidden defects in grocery store items, incomplete information in provider–patient interactions, or restricted access to locked devices). SimpleToM contains concise stories (e.g., "The can of Pringles has moldy chips in it. Mary picks up the can in the supermarket and walks to the cashier."), each with three questions that test different degrees of ToM reasoning, asking models to predict: (a) mental states ("Is Mary aware of the mold?"), (b) behaviors ("Will Mary pay for the chips or report the mold?"), and (c) judgments ("Mary paid for the chips. Was that reasonable?").

A dataset of simulated dialogues is proposed in[454] from the Cooperative Remote Search Task (CReST) corpus.

Uniquely among LLM-orientated benchmarks, **DynToM** generates, through a systematic four-step framework, 1,100 social contexts encompassing 5,500 scenarios and 78,100 questions, each validated for realism and quality, in order to test LLMs' ability to understand and track the temporal progression of mental states across interconnected scenarios. Their average performance underperforms humans by 44.7%, with performance degrading significantly when tracking and reasoning about the shift of mental states.

### 6.1.2 Video and Multimodal

Among datasets encompassing video, image or (more in general) multimodal data, **BDIQA**[455] is a new benchmark in which video question answering is used to explore cognitive reasoning through Theory of Mind. Related to this, an original ToM benchmark from egocentric videos (**Egotom**) is proposed in[456].

MovieGraph-ToM[457] interestingly proposes a *data construction* pipeline from raw movie data to a final, graph-based query answering dataset. The resulting **MovieGraph-ToM** benchmark is a large-scale, deeply annotated dataset built upon a corpus of 30 feature-length films with 65,600 questions designed to probe deep, hierarchical social reasoning.

Multimedia *video large language models* (VLLMs)[458] are considerd in[459], which introduces a new frame localisation benchmark, Theory of Mind Localisation (**ToMLoc**), and shows that fine-tuning a state-of-the-art Video-ChatGPT model on ToMLoc significantly improves performance on Social-IQ 2.0.

**MOMENTS**[460] (Multimodal Mental States) is a comprehensive benchmark designed to assess the ToM capabilities of *multimodal large language models* (MLLMs)[461,462] through realistic, narrative-rich scenarios presented in short films. MOMENTS includes over 2,300 multiple-choice questions spanning seven distinct ToM categories. The benchmark features long video context windows and realistic social interactions that provide deeper insight into characters' mental states. The authors evaluate several MLLMs and find that although vision generally improves performance, models still struggle to integrate it effectively.

**SoMiToM**[463] constructs a challenging dataset containing 35 third-person perspective videos, 363 first-person perspective images, and 1,225 expert-annotated multiple-choice questions (three options). On this dataset, the authors systematically evaluate the performance of human subjects and several state-of-the-art large vision-language models (LVLMs), showing their inferior performance to humans'.

Within the robotics domain, **ToMCAT**[464] is a new benchmark for socially assistive robots with Theory of Mind of children assembling "tangram" puzzles[465], derived from videos of children building tangram puzzles while being assisted by a social robot.

**MMToM-QA** is a benchmark designed to evaluate machine ToM both on multimodal data and on different kinds of unimodal data about a person's activity in a household environment. Each question is associated with a video stream and text input. To confidently choose the correct answer, a model must fuse relevant information from both the video and the text.

Last but not least, **MumaToM**[466] is the first multi-modal Theory of Mind benchmark that evaluates mental reasoning in embodied multi-agent interactions. Videos and text descriptions of people's multi-modal behavior in realistic household environments are provided; based on the context, questions are then asked about people's goals, beliefs, and beliefs about others' goals.

### 6.1.3 Other data

Possibly the most interesting class of new benchmarks comprises original sources of data that go beyond text or images/videos.

**MindCraft**[467], for instance, introduces a fine-grained dataset of collaborative tasks performed by pairs of human subjects in the 3D virtual blocks world of Minecraft. It provides information that captures partners' beliefs of the world and of each other as an interaction unfolds, bringing abundant opportunities to study human collaborative behaviors in situated language communication.

In[468], a tailored testbed game in which a player competes against an agent with a *frustration model* based on theory is created by collecting gameplay data and annotated ground truth about the player's appraisal of the agent's frustration. Face recognition is applied to estimate the player's emotional state.

**ToM-SSI**[469] attempts to go beyond variations of the Sally-Anne test (which only offers a very limited perspective on ToM and neglects the complexity of human social interactions) to specifically test ToM capabilities in environments rich with social interactions and spatial dynamics. ToM-SSI is multimodal and includes group interactions of up to four agents that communicate and move in situated environments, allowing the study of mixed cooperative-obstructive settings and reasoning about multiple agents' mental state in parallel.

Also quite originally, **COKE** formalizes ToM as a collection of 45k+ manually verified *cognitive chains*[470] that characterize human mental activities and subsequent behavioral/affective responses when facing specific social circumstances, using five types of nodes for those cognitive chains: situations, clues, thoughts, actions and emotions.

Within the domain of human-AI interaction, a significant effort is[471], where the authors started with an initial total dataset of 1,513 responses from six surveys (S0 to S5) and consolidated them into a dataset of 1,132 all valid, complete responses from students who indicated that they interacted with JW (an ML question-answering conversational agent).

### 6.1.4 Benchmarks and scoring frameworks

Other effors have focused on "benchmarks", rather than actual datasets, or scoring frameworks.

**COMMON-TOM**[472] is a benchmark which attempts to establishe a common ground between my own view and those of others, arguably the first ToM dataset based on naturally occurring spoken dialogs, using a corpus of 1,710 turns of dyadic dialog across four conversations from CALLHOME[473], and annotate them for common ground.

Perspective taking and its influence is the topic of[474]. There, a **SimToM** benchmark and scoring system are proposed in which a novel two-stage prompting framework inspired by Simulation Theory's notion of perspective-taking is adopted by first filtering context based on what the character in question knows before answering a question about their mental state.

Interestingly, an automated scoring tool coming with a small dataset, called **TrAIngles**, is devised in[475], developed through the fine-tuning of large language models. TrAIngles is trained on a dataset of N ~ 11,000 responses collected from 581 Italian-speaking participants aged between 8 and 81 years.

**TactfulToM**[476] is a novel English benchmark designed to evaluate LLMs' ability to understand white lies within real-life conversations and reason about "prosocial" motivations behind them, generated through a multi-stage *human-in-the-loop*[477] pipeline where LLMs expand manually designed seed stories into conversations to maintain the information asymmetry between participants necessary for authentic white lies.

An innovative benchmark suite of computational theory of mind tasks called **Standoff**, based on the competitive feeding paradigm from comparative psychology can be found in[478].

Finally, **ExploreToM**[479] is the first framework to allow large-scale generation of diverse and challenging theory of mind data for robust training and evaluation. The approach leverages an A* search over a custom domain-specific language to produce complex story structures and novel, diverse, yet plausible scenarios to stress test the limits of large language models.

### 6.1.5 Summary of recent datasets and benchmarks

While extensive, this survey cannot claim to be exaustive, given the rapid pace of evolution of the field – however, it does provide a pretty comprehensive overview of capabilities and limitations of the most recent machine ToM benchmarks.

**Table 1**: *Machine ToM benchmarks classified in terms of nature of the data, etc*.

| | Bench/ Dataset | MM | Data | Tasks | ML models | Size | Year |
|---|---|---|---|---|---|---|---|
| *MMToM-QA* | D | √ | Video + text | True/false belief, belief tracking, goal inference | BIP-ALM, LLMs, LMMs | 134 videos, 600 queries | 2024 |
| *COKE* | D | × | Knowledge graphs | Cognitive reasoning | LLMs | 45K chains | 2024 |
| *FANToM* | D | × | Verbal | 6 different tasks | LLMs | 10K queries, 256 convers. | 2023 |
| *BDIQA* | D | × | Video | Video QA, BDI reasoning | Memory-based, Hierarchical Module, GNN and pretrained models (ClipBERT, Frozen, JustAsk) | 3,527 videos, 19,932 QA pairs | 2024 |
| *OpenToM* | D | × | Verbal | Questions | LLMs | 696 narratives | 2024 |
| | | | | (location, attitude, multi-hop) as binary or ternary classification tasks | | gen via GPT-3.5/4-Turbo | |
| *ToMChallenges* | D | × | Verbal | Sally-Anne test, Smarties test | text-davinci-003 and gpt-3.5-turbo-0301, and gpt-4-0613 | 30 variations of each test | 2023 |
| *ToMBench* | D | × | Verbal | 8 tasks, 31 abilities | 10 LLMs | 2,860 samples | 2024 |
| *Multi-ToM* | D | × | Verbal, multi-language | ? | ? | ? | 2024 |
| *EPITOME* | D | × | Verbal | False belief, rec. mindreading,short/ strange stories, indirect request, scalar implicature | LLMs | 6 experiments, 613 trials | 2023 |
| *EgoToM* | D | √ | Video | QA | MLLMs (Llama3.1-instruct, GPT, VLMs) | 7k 5-min clips from Ego4D; QA gen bia GPT-4-Turbo | 2025 |
| *SUITE* | D | × | Verbal | Standard, no context, shuffle; multiple choice questions | LLMs (Qwen-Flash, Qwen-Plus, LLaMA-3.3-70B) | 38 users, >6,000 words per user | 2026 |
| *Hi-ToM* | D | × | Verbal | Prompting (Vanilla and Chain-of-Thought) | LLMs (GPT-3.5-Turbo, Claude-instant, Guanaco) | ? | 2023 |
| *MovieGraph-ToM* | D | √ | Video + text | ToM, behaviour, social inference tasks | DeepSeek, GPT, Gemini, Llava, Mistral, InternVL | 30 films, 100k ToM states, 9k causal links, 48k nodes; 65,600 queries | 2026 |
| *ToMLoc* | B | × | Video from Social-IQ 2.0 dataset | Frame localisation and question answering | Video-ChatGPT + ActivityNet/ToMLoc, Leave-Frame-Out | 1,403 video clips, 8,076 questions | 2025 |
| *MindCraft* | D | √ | Virtual world + text | Pairs of users to create materials in Minecraft; queries on completed task, player knowledge | Various computational models | 100 games | 2021 |
| *SoMiToM* | D | √ | Video + text | First-person perspective, third-person perspective | large vision-language models | 35 3rd-person videos, 363 1st-person images; 1225 questions | 2025 |
| *RecToM* | D | × | Verbal | Belief, desire, judgement, | Deepseek-V3, GPT-4o-mini, | 253 dialogues from the | |

| | | | | reasoning | Gemini 2.5 Flash-Lite, Claude 3.5 | ReDial dataset | |
|---|---|---|---|---|---|---|---|
| *AnaToM* | B | × | Verbal | Anatomy of difficulty in LLM reasoning | LLMs | 1,000 instances | 2025 |
| *ToMCAT* | D | ? | Videos of child-robot interactions | Tangram puzzles of various levels of difficulty | Claude Sonnet 3.5, no fine-tuning | 436 observations of 34 children | 2025 |
| *COMMON-TOM* | B | × | Verbal, from English-only CALLHOME corpus | Classification of agent belief states assuming common ground | Various LLMs (GPT-3.5, GPT-4, Mistral-7B-Instruct) | 1,710 turns of dialog across 4 conversations 7,374 queries | 2024 |
| *SimToM* | B/F | × | Prompting framework | False belief | LLMs | ? | 2024 |
| *UniToMBench* | D | × | Verbal | Multi-interaction, evolving stories, | GPT-4o, GPT-40 Mini | 3-5 narrative per task | 2025 |
| *MOMENTS* | D | √ | Videos from SF20K dataset | 7 ToM categories | MLLMs (video, audiovisual and speech only) | 2,300 multiple-choice queries | 2025 |
| *PersuasiveToM* | D | × | Verbal, from Daily Persuasion dataset | ToM reasoning (belief, desire, intention) and application (persuation, judgement) | LLMs | Questions from GPT-4o; 35 domains, 525 dialogues, ca 11.5k questions | 2025 |
| *DialToM* | D | × | Verbal, from AnnoMI, ESConv, PFG | Retrospective inf., prospective diagnostic forecast | Various LLMs | 5,943 contexts, ca 42k questions | 2026 |
| *ToM-SSI* | D | √ | Group interactions in a grid | Percept, beliefs, intentions | 15 LLMs and VLMs | 6,000 questions, 5 tasks | 2025 |
| *DebaToM* | B | × | Verbal, from BigTom and FanToM | Forward belief, forward action, backward belief | CoT, SimToM, TIMEToM, CCoToM, EnigmaToM | × | 2026 |
| *CogToM* | D | × | Verbal (bilingual), debates | 36 tasks across Emotion, Desire, Intention, Percept, Knowledge, Belief and Non-literal categories | 22 models, including GPT-5.1 and Qwen3-Max | 8000 bilingual instances, 46 paradigms, validated by 49 human annotators | 2026 |
| *TrAIngles* | S | × | Verbal | F1 score | 5 Italian BERT models | 11k responses, 581 Italian speakers | 2026 |
| *TOMATO* | D | × | Verbal | Multiple choice QA; false belief detection | 9 LLMs; Llama-3-70B-Instruct for generation | 800 conversations | 2025 |
| *TactfulToM* | B | × | Verbal, English | White lie understanding, | LLMs; GPT-4o for generation | 100 conversations, | 2025 |

| | | | | white lie reasoning | | 6.7k questions | |
|---|---|---|---|---|---|---|---|
| *Standoff* | B | × | Simulated, non-verbal | Accuracy of dominant opponent's belief | Convolutional LSTM | × | 2024 |
| *ExploreToM* | B | × | A* algo for adversarially constructed stories | Accuracy on false belief datasets (ToMi, Hi-ToM, BigToM, OpenToM, FANToM) | LLMs (Llama-3.1-70B-Instruct, GPT-4o, Mixtral-8x7B-Instruct) | 10 story structures for 9 action sets | 2025 |
| *BigToM* | D | × | Verbal | Beliefs, percepts | LLMs | 5,000 evaluations | 2023 |
| *SimpleToM* | D | × | Verbal | Mental state, behaviour, judgement | Various LLMs; Claude-3-Opus, GPT-4 for gener. | 3,600 stories | 2024 |
| *Muma-ToM* | D | √ | Video + questions | Belief, social goals, etc | VLMs, LLMs (GPT-4o, Llava 1.6) and BIP-ALM | 18 participants, 90 questions | 2025 |
| *DynToM* | D | × | Verbal | Understanding, transformation questions | 10 LLMs, including GPT-4o, Mixtral-8x7B, DeepSeek-V2 | 1,100 contexts 5,500 scenarios, 78k questiones | 2025 |

## 6.2 Benchmarking the benchmarks

Let us conclude by reviewing the above benchmarks from the point of view of their ability to elicit Principles 1-8. Overall, as all of these are static benchmarks, none of them satisfies Principle 5, in that they do not contemplate any active learning process. For the same reason, they do not incorporate Principle 1, i.e., the need for machine ToM to continually update itself. The number of continual learning datasets in machine learning is growing very quickly (see, e.g.[480,481,482]), but it remains focused on simple classification or visual recognition tasks. Similarly, the modelling and quantification of uncertainty (Principle 8) is widely ignored by the existing battery of datasets and scoring frameworks[483]. Higher-order ToM (Principle 6) is contemplated by at least one dataset: Hi-ToM. Principle 7 (theory of mind as a collection of diverse tasks) is instead widely recognized by pretty much all the datasets reviewed above, even the simplest; yet, tasks are generally evaluated separately, rather than holistically as a whole.
Principle 2, i.e., the role of the self-simulation, is quite transparent to the data used – therefore, models employing it can be applied to solving any of the tasks/situations contemplated in current benchmarks. The same holds for Principle 4 (combining simulation-theory and theory-theory), as the internal engine of the ToM model has little to do with the nature of the tests/data. In practice, however, the choice of baseline models in the associated papers is quite restriticve: this is certainly true of datasets focused on ToM abilities in LLMs (the vast majority), where no other classes of machine ToM approaches is considered. Exceptions in this sense are BDIQA, MMToM-QA, MindCraft, DebaToM and (partially) MOMENTS.
Principle 3 is probably the trickiest. Inverse reinforcement learning strategies are normally tested in the literature using dedicated datasets which have only tangentially to do with Theory of Mind[484]. An exception seems to be PsychSim[485]; yet, the disconnect with IRL and ToM in terms of benchmarking is clear and something to be addressed in the future within the proposed holistic framework.

# 7 An agenda for the future

Other recent position papers have recently attempted to suggest a suitable research agenda for the future of machine theory of mind[486,487,488,489]. None, however, has done this through the lense of a set of evidence-backed Principles informing, in turn, a holistic meta-model through which all existing efforts can be read and interpreted.

Our holistic approach to machine theory of mind does raise one important question: can one model non-ToM-directed actions in a self-contained computational ToM model? In other words, can one design a ToM module for an Observer without considering its wider purposes? Past efforts have mostly been focusing on the Theory of Mind component, thus neglecting its role as just one element of larger entity for which ToM is not necessarily the main goal.

From our analysis, a number of clear messages emerge: (1) while partly acknowledged, the role of continual learning in Machine ToM is still neglected in practice, and will need to drive the evolution of the field in the future; (2) few of the proposed methods actually put simulation-theory and theory-theory together; (3) from the point of the view of where the clues for ToM come from, the notion that observation and interaction should be seemlessly intertwined with, as an imitation of humans' own lifelong learning experience, has been arguably severely overlooked; (4) the same holds for the role of active learning, while (5) uncertainty quantification, while partly addressed by Bayesian ToM methods, has not progressed to model epistemic uncertainty in a systematic treatment of all the components of a machine ToM.

It seems to us that, a proper treatment of theory of mind that is not partial or blind to important aspects of the problem needs to start from a holistic perspective of the Observer agent, of which ToM is only one aspect, albeit important. It needs to recognise the continual learning under uncertainty nature of the problem, and the role of knowledge and data in this process. It should also probably refrain from empirically analysing black box models' ability to solve (mostly verbal) ToM tasks, something that, in our mind, adds little to our understanding of the problem. Multimodality is also barely studied, whereas humans (the only creatures, so far, certain to possess theory of mind) do explore the world using multiple senses, of which vision is one of the most important.

From an interdisciplinary perspective, more needs to be done to bring together the many communities interested in the problem, if we are to avoid reinventing the wheel and, most of all, we wish to advance towards a general theory of Machine Theory of Mind. Finally, much work will need to be done on the computational side to render our abstract meta-model as actual computational engines, e.g. by exploring the potential of architecture search and generative techniques for ST/TT, something that will likely stimulate advances within meta- and automated machine learning themselves.

[9] M. Dastani, 2APL: A practical agent programming language, Autonomous Agents and Multi-agent Systems 16(3) (2008), 214–248
[10] Shanton, Karen, and Alvin Goldman. "Simulation theory." *Wiley Interdisciplinary Reviews: Cognitive Science* 1.4 (2010): 527-538.
[11] Harris, P. L. (1992). From simulation to folk psychology: The case for development. *Mind & Language*, 7(1-2), 120-144
[12] Scholl, B. J., and Leslie, A. M. (1999). Modularity, development and 'theory of mind'. *Mind & Language*, 14(1), 131-153
[13] Harbers, Maaike, Karel Van den Bosch, and John-Jules Meyer. "Modeling agents with a theory of mind: Theory–theory versus simulation theory." *Web Intelligence and Agent Systems* 10.3 (2012): 331-343.
[14] Wagner, Eitan, et al. "Mind your theory: Theory of mind goes deeper than reasoning." *Findings of the Association for Computational Linguistics: ACL 2025*. 2025.
[15] Saxe, Rebecca, and Simon Baron-Cohen. "The neuroscience of theory of mind." (2006): 1-9.
[16] Apperly, Ian A. "Beyond Simulation–Theory and Theory–Theory: Why social cognitive neuroscience should use its own concepts to study "theory of mind"." *Cognition* 107.1 (2008): 266-283.
[17] Gallese, Vittorio, and Alvin Goldman. "Mirror neurons and the simulation theory of mind-reading." *Trends in cognitive sciences* 2.12 (1998): 493-501.
[18] Schulte-Rüther, Martin, et al. "Mirror neuron and theory of mind mechanisms involved in face-to-face interactions: a functional magnetic resonance imaging approach to empathy." *Journal of cognitive neuroscience* 19.8 (2007): 1354-1372.
[19] Borg, Emma. "If mirror neurons are the answer, what was the question?." *Journal of Consciousness Studies* 14.8 (2007): 5-19.
[20] Heyes, Cecilia, and Caroline Catmur. "What happened to mirror neurons?." *Perspectives on Psychological Science* 17.1 (2022): 153-168.
[21] Mahy, Caitlin EV, Louis J. Moses, and Jennifer H. Pfeifer. "How and where: Theory-of-mind in the brain." *Developmental cognitive neuroscience* 9 (2014): 68-81.
[22] Carrington, Sarah J., and Anthony J. Bailey. "Are there theory of mind regions in the brain? A review of the neuroimaging literature." *Human brain mapping* 30.8 (2009): 2313-2335.
[23] Tomasello, Rosario, Isabella Boux, and Friedemann Pulvermüller. "Theory of Mind and the brain substrates of direct and indirect communicative action understanding." *Philosophical Transactions B* 380.1932 (2025): 20230497.
[24] Yin, Xiaoyun, et al. "When Researchers Say Mental Model/Theory of Mind of AI, What Are They Really Talking About?." *arXiv preprint arXiv:2510.02660* (2025).
[25] Matta, Daoud. "Artificial intelligence and theory of mind." *Journal of Psychology and AI* 2.1 (2026): 2628373.
[26] Garcia-Lopez, Alvaro. "Theory of mind in artificial intelligence applications." *The theory of mind under scrutiny: Psychopathology, neuroscience, philosophy of mind and artificial intelligence*. Cham: Springer Nature Switzerland, 2024. 723-750.
[27] Gerrans, Philip. "The theory of mind module in evolutionary psychology." *Biology and Philosophy* 17.3 (2002): 305-321.
[28] Apperly, Ian. *Mindreaders: the cognitive basis of" theory of mind"*. Psychology Press, 2010.
[29] Apperly, Ian A. "What is "theory of mind"? Concepts, cognitive processes and individual differences." *Quarterly journal of experimental psychology* 65.5 (2012): 825-839.
[30] Langley, Christelle, et al. "Theory of mind and preference learning at the interface of cognitive science, neuroscience, and AI: A review." *Frontiers in artificial intelligence* 5 (2022): 778852.
[31] Langley, Christelle, et al. "Theory of Mind in Humans and in Machines." *Frontiers in artificial intelligence* 5 (2022): 917565.
[32] Langley, Christelle, Fabio Cuzzolin, and Barbara J. Sahakian. "Neuroscience for AI: The importance of theory of mind." *Developments in Neuroethics and Bioethics*. Vol. 7. Academic Press, 2024. 65-83.
[33] Yıldız, Tolga. "The minds we make: A philosophical inquiry into theory of mind and artificial intelligence." *Integrative Psychological and Behavioral Science* 59.1 (2025): 10.
[34] Bzdok, Danilo, et al. "Parsing the neural correlates of moral cognition: ALE meta-analysis on morality, theory of mind, and empathy." *Brain Structure and Function* 217.4 (2012): 783-796.
[35] Lagattuta, Kristin Hansen, and Drika Weller. "Interrelations between theory of mind and morality: A developmental perspective." *Handbook of moral development*. Psychology Press, 2013. 385-407.
[36] Luhrmann, Tanya, et al. "Toward an anthropological theory of mind." *Suomen Antropologi: Journal of the Finnish Anthropological Society* 36.4 (2011): 5-69.
[37] Eisenmajer, Richard, and Margot Prior. "Cognitive linguistic correlates of 'theory of mind'ability in autistic children." *British journal of developmental psychology* 9.2 (1991): 351-364.
[38] De Villiers, Jill. "The interface of language and theory of mind." *Lingua* 117.11 (2007): 1858-1878.
[39] Dixson, Henry GW, et al. "Scaling theory of mind in a small-scale society: A case study from Vanuatu." *Child Development* 89.6 (2018): 2157-2175.
[40] Krupenye, Christopher, and Josep Call. "Theory of mind in animals: Current and future directions." *Wiley Interdisciplinary Reviews: Cognitive Science* 10.6 (2019): e1503.
[41] Heyes, Cecilia M. "Theory of mind in nonhuman primates." *Behavioral and brain sciences* 21.1 (1998): 101-114.
[42] Bianco, Francesca, and Dimitri Ognibene. "Functional advantages of an adaptive theory of mind for robotics: a review of current architectures." *2019 11th Computer Science and Electronic Engineering (CEEC)* (2019): 139-143.
[43] Çelikok, Mustafa Mert, et al. "Interactive AI with a theory of mind." *arXiv preprint arXiv:1912.05284* (2019).

[44] Alexopoulos, George S. "Loneliness, theory of mind, and artificial intelligence." *The American Journal of Geriatric Psychiatry* (2025).
[45] Harré, Michael S., Catherine Drysdale, and Jaime Ruiz-Serra. "An AI Theory of Mind Will Enhance Our Collective Intelligence." *arXiv preprint arXiv:2411.09168* (2024).
[46] Schneider, Howard. "Theory of Mind as a Core Component of Artificial General Intelligence." *International Conference on Artificial General Intelligence*. Cham: Springer Nature Switzerland, 2025.
[47] Zilberman, Jacob, and Boaz Tamir. "AGI imagined: how is AGI configured by the theories of mind." *AI & SOCIETY* (2025): 1-21.
[48] Schossau, Jory, and Arend Hintze. "Towards a theory of mind for artificial intelligence agents." *Artificial Life Conference Proceedings 35*. Vol. 2023. No. 1. One Rogers Street, Cambridge, MA 02142-1209, USA journals-info@ mit. edu: MIT Press, 2023.
[49] Yıldız, Tolga. "The minds we make: A philosophical inquiry into theory of mind and artificial intelligence." *Integrative Psychological and Behavioral Science* 59.1 (2025): 10.
[50] Lopez-Soto, Teresa. *The theory of mind under scrutiny: Psychopathology, neuroscience, philosophy of mind and artificial intelligence*. Eds. Alvaro Garcia-Lopez, and Francisco J. Salguero-Lamillar. Springer Nature, 2024.
[51] Gurney, Nikolos, and David V. Pynadath. "Robots with theory of mind for humans: A survey." *2022 31st IEEE International Conference on Robot and Human Interactive Communication (RO-MAN)*. IEEE, 2022.
[52] Chen, Boyuan, Carl Vondrick, and Hod Lipson. "Visual behavior modelling for robotic theory of mind." *Scientific Reports* 11.1 (2021): 424.
[53] Scassellati, Brian. "Theory of mind for a humanoid robot." *Autonomous Robots* 12.1 (2002): 13-24.
[54] He, Chengyang, et al. "Latent theory of mind: A decentralized diffusion architecture for cooperative manipulation." *arXiv preprint arXiv:2505.09144* (2025).
[55] Chandra, Rohan, Aniket Bera, and Dinesh Manocha. "Stylepredict: Machine theory of mind for human driver behavior from trajectories." *arXiv preprint arXiv:2011.04816* (2020).
[56] Montese, Sara. *Policy graphs and theory of mind for explainable autonomous driving*. Diss. Politecnico di Torino, 2024.
[57] Nori, Raffaella, et al. "The prevention of road accidents in non-expert drivers: Exploring the influence of Theory of Mind and driving style." *Safety science* 175 (2024): 106516.
[58] Langley, Christelle, et al. "Theory of mind and preference learning at the interface of cognitive science, neuroscience, and AI: A review." *Frontiers in artificial intelligence* 5 (2022): 778852.
[59] Mao, Yuanyuan, et al. "A review on machine theory of mind." *IEEE Transactions on Computational Social Systems* 11.6 (2024): 7114-7132.
[60] Nebreda, Alberto, et al. "The social machine: artificial intelligence (AI) approaches to theory of mind." *The theory of mind under scrutiny: psychopathology, neuroscience, philosophy of mind and artificial intelligence*. Cham: Springer Nature Switzerland, 2024. 681-722.
[61] Colombo, Matteo, and Gualtiero Piccinini. *The computational theory of mind*. Cambridge University Press, 2023.
[62] Sclar, Melanie, Graham Neubig, and Yonatan Bisk. "Symmetric machine theory of mind." *International conference on machine learning*. PMLR, 2022.
[63] Li, Huao, et al. "Theory of mind for multi-agent collaboration via large language models." *Proceedings of the 2023 Conference on Empirical Methods in Natural Language Processing*. 2023.
[64] Canese, Lorenzo, et al. "Multi-agent reinforcement learning: A review of challenges and applications." *Applied Sciences* 11.11 (2021): 4948.
[65] Wang, Lingyi, et al. "MetaMind: General and Cognitive World Models in Multi-Agent Systems by Meta-Theory of Mind." *arXiv preprint arXiv:2603.00808* (2026).
[66] Cross, Logan, et al. "Hypothetical minds: Scaffolding theory of mind for multi-agent tasks with large language models." *International Conference on Learning Representations*. Vol. 2025. 2025.
[67] Khan, Shakil M., and Maryam Rostamigiv. "On Explaining Agent Behaviour via Root Cause Analysis: A Formal Account Grounded in Theory of Mind." *ECAI*. 2023.
[68] Akula, Arjun R., et al. "X-tom: Explaining with theory-of-mind for gaining justified human trust." *arXiv preprint arXiv:1909.06907* (2019).
[69] Wu, Jincenzi, et al. "Coke: A cognitive knowledge graph for machine theory of mind." *Proceedings of the 62nd Annual Meeting of the Association for Computational Linguistics (Volume 1: Long Papers)*. 2024.
[70] Jin, Chuanyang, et al. "Mmtom-qa: Multimodal theory of mind question answering." *Proceedings of the 62nd Annual Meeting of the Association for Computational Linguistics (Volume 1: Long Papers)*. 2024.
[71] Sclar, Melanie, et al. "Explore theory of mind: Program-guided adversarial data generation for theory of mind reasoning." *International Conference on Learning Representations*. Vol. 2025. 2025.
[72] Zhu, Fengming, et al. "A Computable Game-Theoretic Framework for Multi-Agent Theory of Mind." *arXiv preprint arXiv:2511.22536* (2025).
[73] Nguyen, Dung, et al. "Memory-augmented theory of mind network." *Proceedings of the AAAI Conference on Artificial Intelligence*. Vol. 37. No. 10. 2023.
[74] Shi, Haojun, et al. "MuMa-ToM: Multi-modal multi-agent theory of mind." *Proceedings of the AAAI Conference on Artificial Intelligence*. Vol. 39. No. 2. 2025.
[75] Bortoletto, Matteo, et al. "Explicit modelling of theory of mind for belief prediction in nonverbal social interactions." *arXiv preprint arXiv:2407.06762* (2024).

[76] Ashktorab, Zahra, et al. "Bridging the Gap: Unifying HCI & ML Perspectives on Mutual Theory of Mind." *International Joint Conference on Artificial Intelligence*. 2025.
[77] Oguntola, Ini, Dana Hughes, and Katia Sycara. "Deep interpretable models of theory of mind." *2021 30th IEEE international conference on robot & human interactive communication (RO-MAN)*. IEEE, 2021.
[78] Erdogan, Emre, et al. "TOMA: computational theory of mind with abstractions for hybrid intelligence." *Journal of Artificial Intelligence Research* 82 (2025): 285-311.
[79] Sarangi, Sneheel, Maha Elgarf, and Hanan Salam. "Decompose-ToM: Enhancing theory of mind reasoning in large language models through simulation and task decomposition." *Proceedings of the 31st international conference on computational linguistics*. 2025.
[80] Williams, Jessica, Stephen M. Fiore, and Florian Jentsch. "Supporting artificial social intelligence with theory of mind." *Frontiers in artificial intelligence* 5 (2022): 750763.
[81] Harré, Michael S., Jaime Ruiz-Serra, and Catherine Drysdale. "Artificial Theory of Mind and Self-Guided Social Organisation." *arXiv preprint arXiv:2411.09169* (2024).
[82] Shapira, Natalie, et al. "Clever hans or neural theory of mind? stress testing social reasoning in large language models." *Proceedings of the 18th Conference of the European Chapter of the Association for Computational Linguistics (Volume 1: Long Papers)*. 2024.
[83] Ho, Mark K., Rebecca Saxe, and Fiery Cushman. "Planning with theory of mind." *Trends in Cognitive Sciences* 26.11 (2022): 959-971.
[84] Xu, Hainiu, et al. "Enigmatom: Improve llms' theory-of-mind reasoning capabilities with neural knowledge base of entity states." *Findings of the Association for Computational Linguistics: ACL 2025*. 2025.
[85] Van Duijn, Max, et al. "Theory of mind in large language models: Examining performance of 11 state-of-the-art models vs. children aged 7-10 on advanced tests." *Proceedings of the 27th conference on computational natural language learning (CoNLL)*. 2023.
[86] Kosinski, Michal. "Evaluating large language models in theory of mind tasks." *Proceedings of the National Academy of Sciences* 121.45 (2024): e2405460121.
[87] Verma, Mudit, Siddhant Bhambri, and Subbarao Kambhampati. "Theory of mind abilities of large language models in human-robot interaction: An illusion?." *Companion of the 2024 ACM/IEEE International Conference on Human-Robot Interaction*. 2024.
[88] Zhou, Pei, et al. "How far are large language models from agents with theory-of-mind?." *arXiv preprint arXiv:2310.03051* (2023).
[89] Amirizaniani, Maryam, et al. "Can llms reason like humans? assessing theory of mind reasoning in llms for open-ended questions." *Proceedings of the 33rd ACM International Conference on Information and Knowledge Management*. 2024.
[90] Sap, Maarten, et al. "Neural theory-of-mind? on the limits of social intelligence in large lms." *Proceedings of the 2022 conference on empirical methods in natural language processing*. 2022.
[91] Ullman, Tomer. "Large language models fail on trivial alterations to theory-of-mind tasks." *arXiv preprint arXiv:2302.08399* (2023).
[92] Sclar, Melanie, et al. "Minding language models'(lack of) theory of mind: A plug-and-play multi-character belief tracker." *Proceedings of the 61st Annual Meeting of the Association for Computational Linguistics (Volume 1: Long Papers)*. 2023.
[93] Ma, Ziqiao, et al. "Towards a holistic landscape of situated theory of mind in large language models." *Findings of the association for computational linguistics: EMNLP 2023*. 2023.
[94] Strachan, James WA, et al. "Testing theory of mind in large language models and humans." *Nature human behaviour* 8.7 (2024): 1285-1295.
[95] Aru, Jaan, et al. "Mind the gap: Challenges of deep learning approaches to theory of mind." *arXiv preprint arXiv:2203.16540* (2022).
[96] Nickel, Christian, Laura Schrewe, and Lucie Flek. "Probing the robustness of theory of mind in large language models." *arXiv preprint arXiv:2410.06271* (2024).
[97] Tang, Weizhi, and Vaishak Belle. "Tom-lm: Delegating theory of mind reasoning to external symbolic executors in large language models." *International Conference on Neural-Symbolic Learning and Reasoning*. Cham: Springer Nature Switzerland, 2024.
[98] Moghaddam, Shima Rahimi, and Christopher J. Honey. "Boosting theory-of-mind performance in large language models via prompting." *arXiv preprint arXiv:2304.11490* (2023).
[99] Wester, Joel, et al. "Theory of mind and self-presentation in human-LLM interactions." *Adjunct Proceedings of the ACM SIGCHI Conference on Human Factors in Computing Systems*. 2024.
[100] Muchovej, John, et al. "GPT-4o Lacks Core Features of Theory of Mind." *arXiv preprint arXiv:2602.12150* (2026).
[101] Pîrjan, Alexandru, and Dana-Mihaela Petroşanu. "Proposing A Multi-Agent Simulation Framework For Quantifying The Emergence Of Theory Of Mind In Foundation Artificial Intelligence Models." *Journal of Information Systems & Operations Management* 19.2 (2025): 364-408.
[102] Sarangi, Sneheel, et al. "Agentic-ToM: Cognition-Inspired Agentic Processing For Enhancing Theory of Mind Reasoning in Large Language Models." *30th Conference on Empirical Methods in Natural Language Processing, EMNLP 2025*. Association for Computational Linguistics (ACL), 2025.
[103] Grislain, Clémence, et al. "Utility-based Adaptive Teaching Strategies using Bayesian Theory of Mind." *arXiv preprint arXiv:2309.17275* (2023).

[104] Kleiman-Weiner, Max, et al. "Evolving general cooperation with a Bayesian theory of mind." *Proceedings of the National Academy of Sciences* 122.25 (2025): e2400993122.
[105] Ying, Lance, et al. "Grounding language about belief in a bayesian theory-of-mind." *arXiv preprint arXiv:2402.10416* (2024).
[106] Zhang, Shanshan, Chuyang Wu, and Jussi PP Jokinen. "Quantifying Uncertainty in Machine Theory of Mind Across Time." *CEUR Workshop Proceedings*. RWTH Aachen, 2024.
[107] Huang, X. Angelo, et al. "A notion of complexity for theory of mind via discrete world models." *Findings of the Association for Computational Linguistics: EMNLP 2024*. 2024.
[108] Manenti, Giorgio L., et al. "BELIEFS: A Hierarchical Theory of Mind Model based on Strategy Inference." *bioRxiv* (2025): 2025-06.
[109] Ayache, Julia, et al. "Humanness lies in unpredictability: Role of Theory of Mind on anthropomorphism in human-computer interactions." *Proceedings of the 10th International Conference on Human-Agent Interaction*. 2022.
[110] Zhang, Chunhui, et al. "Scaling Multimodal Theory-of-Mind with Weak-to-Strong Bayesian Reasoning." (2025).
[111] Zilberman, Jacob, and Boaz Tamir. "AGI imagined: how is AGI configured by the theories of mind." *AI & SOCIETY* (2025): 1-21.
[112] Zilberman, Jacob, and Boaz Tamir. "AGI imagined: how is AGI configured by the theories of mind." *AI & SOCIETY* (2025): 1-21.
[113] Hendrycks, Dan, et al. "A definition of AGI." *arXiv preprint arXiv:2510.18212* (2025).
[114] Schneider, Howard. "Theory of Mind as a Core Component of Artificial General Intelligence." *International Conference on Artificial General Intelligence*. Cham: Springer Nature Switzerland, 2025.
[115] Huang, Ming-Hui, and Roland T. Rust. "The caring machine: Feeling AI for customer care." *Journal of Marketing* 88.5 (2024): 1-23.
[116] Marchetti, Antonella, et al. "Artificial intelligence and the illusion of understanding: A systematic review of theory of mind and large language models." *Cyberpsychology, Behavior, and Social Networking* 28.7 (2025): 505-514.
[117] y Arcas, Blaise Agüera. "Do large language models understand us?." *Daedalus* 151.2 (2022): 183-197.
[118] Kumar, Prabhat, Adrienne Raglin, and John Richardson. "General agent theory of mind: Preliminary investigations and vision." *International Conference on Human-Computer Interaction*. Cham: Springer Nature Switzerland, 2023.
[119] Pafla, Marvin, et al. "The Need for a Common Ground: Ontological Guidelines for a Mutual Human-AI Theory of Mind."
[120] Krasnytskyi, Nikita, and Fabio Cuzzolin. "Evaluating Machine Theory of Mind: A Critical Analysis of ToMnet-N." *ToM4AI 2025* 4 (2025): 60.
[121] N. C. Rabinowitz, F. Perbet, H. F. Song, C. Zhang, S. Eslami, and M. Botvinick, "Machine theory of mind," in Proc. Int. Conf. Machine Learn., 2018, pp. 4218–4227.
[122] Rudenko, Andrey, et al. "Human motion trajectory prediction: A survey." *The International Journal of Robotics Research* 39.8 (2020): 895-935.
[123] Holzinger, Andreas, et al. "Causability and explainability of artificial intelligence in medicine." *Wiley interdisciplinary reviews: data mining and knowledge discovery* 9.4 (2019): e1312.
[124] Hewson, Joshua TS. "Combining Theory of Mind and Kindness for Self-Supervised Human-AI Alignment." *arXiv preprint arXiv:2411.04127* (2024).
[125] Street, Winnie. "Llm theory of mind and alignment: Opportunities and risks." *arXiv preprint arXiv:2405.08154* (2024).
[126] Love, Miira. "The Role of Theory of Mind in Self-Reflective AI Agents: Enabling Empathy and Value Alignment in Human-AI Collaboration." (2025).
[127] Langley, Christelle, et al. "Theory of mind and preference learning at the interface of cognitive science, neuroscience, and AI: A review." *Frontiers in artificial intelligence* 5 (2022): 778852.
[128] Sullivan, Yulia W., and Samuel Fosso Wamba. "Moral judgments in the age of artificial intelligence." *Journal of Business Ethics* 178.4 (2022): 917-943.
[129] Aoshima, Tatsuhiro, and Mitsuaki Akiyama. "Towards Safety Evaluations of Theory of Mind in Large Language Models." *IEICE Transactions on Information and Systems* (2025).
[130] Brandone, Amanda C., and Wyntre Stout. "The origins of theory of mind in infant social cognition: Investigating longitudinal pathways from intention understanding and joint attention to preschool theory of mind." *Journal of Cognition and Development* 24.3 (2023): 375-396.
[131] Victoria, Bamicha, and Athanasios Drigas. "ToM & ASD: The interconnection of Theory of Mind with the social-emotional, cognitive development of children with Autism Spectrum Disorder. The use of ICTs as an alternative form of intervention in ASD." *Technium Soc. Sci. J.* 33 (2022): 42.
[132] Mao, Yuanyuan, et al. "A review on machine theory of mind." *IEEE Transactions on Computational Social Systems* 11.6 (2024): 7114-7132.
[133] Bamicha, Victoria, and Athanasios Drigas. "The Evolutionary Course of Theory of Mind-Factors that facilitate or inhibit its operation & the role of ICTs." *Technium Soc. Sci. J.* 30 (2022): 138.
[134] Wellman, Henry M. "Developing a theory of mind." *The Wiley-Blackwell handbook of childhood cognitive development* 2 (2011): 258-284.
[135] Rakoczy, Hannes. "Foundations of theory of mind and its development in early childhood." *Nature Reviews Psychology* 1.4 (2022): 223-235.
[136] Nguyen, Thuy Ngoc, and Cleotilde Gonzalez. "Theory of mind from observation in cognitive models and humans." *Topics in cognitive science* 14.4 (2022): 665-686.

[137] Dutemple, Elizabeth, Hanifa Hakimi, and Diane Poulin-Dubois. "Do I know what they know? Linking metacognition, theory of mind, and selective social learning." *Journal of experimental child psychology* 227 (2023): 105572.
[138] Oguntola, Ini, et al. "Theory of mind as intrinsic motivation for multi-agent reinforcement learning." *arXiv preprint arXiv:2307.01158* (2023).
[139] Lică, Mircea, et al. "Mindforge: Empowering embodied agents with theory of mind for lifelong cultural learning." *Advances in Neural Information Processing Systems* 38 (2026): 170024-170053.
[140] Zhinin-Vera, Luis, et al. "Theory of mind and continual reinforcement learning for bullying intervention." *Neural Computing and Applications* 38.6 (2026): 160.
[141] Xiao, Yang, et al. "Towards dynamic theory of mind: Evaluating LLM adaptation to temporal evolution of human states." *Proceedings of the 63rd Annual Meeting of the Association for Computational Linguistics (Volume 1: Long Papers)*. 2025.
[142] Calmette, Tony, and Hélène Meunier. "Is self-awareness necessary to have a theory of mind?." *Biological Reviews* 99.5 (2024): 1736-1771.
[143] Grazzani, Ilaria. "Promoting theory of mind and emotion understanding in preschool settings: an exploratory training study." *Frontiers in Psychology* 15 (2024): 1439824.
[144] Smogorzewska, Joanna, et al. "Just listen to your mind: Consequences of theory of mind development for deaf or hard-of-hearing children." *Research in Developmental Disabilities* 127 (2022): 104261.
[145] Fan, Xianzhe, et al. "SoMi-ToM: Evaluating Multi-Perspective Theory of Mind in Embodied Social Interactions." *Advances in Neural Information Processing Systems* 38 (2026).
[146] Söderlund, Magnus. "Human employees and service robots in the service encounter and the role of attribution of theory of mind." *Journal of Retailing and Consumer Services* 81 (2024): 103964.
[147] Soubki, Adil, et al. "Views are my own, but also yours: Benchmarking theory of mind using common ground." *Findings of the Association for Computational Linguistics: ACL 2024*. 2024.
[148] Liu, Andy, et al. "Computational language acquisition with theory of mind." *arXiv preprint arXiv:2303.01502* (2023).
[149] Zhinin-Vera, Luis, et al. "Mindful human digital twins: integrating theory of mind with multi-agent reinforcement learning." *Applied Soft Computing* 174 (2025): 112939.
[150] Nguyen, Loc X., et al. "A contemporary survey on semantic communications: Theory of mind, generative ai, and deep joint source-channel coding." *IEEE Communications Surveys & Tutorials* (2025).
[151] Aru, Jaan, et al. "Mind the gap: Challenges of deep learning approaches to theory of mind." *arXiv preprint arXiv:2203.16540* (2022).
[152] Jara-Ettinger, Julian. "Theory of mind as inverse reinforcement learning." *Current Opinion in Behavioral Sciences* 29 (2019): 105-110.
[153] Ruiz-Serra, Jaime, and Michael S. Harré. "Inverse reinforcement learning as the algorithmic basis for theory of mind: current methods and open problems." *Algorithms* 16.2 (2023): 68.
[154] Kopp, Stefan, and Nicole Krämer. "Revisiting human-agent communication: The importance of joint co-construction and understanding mental states." *Frontiers in Psychology* 12 (2021): 580955.
[155] Becchio, Cristina, et al. "Seeing mental states: An experimental strategy for measuring the observability of other minds." *Physics of life reviews* 24 (2018): 67-80.
[156] Kaelbling, Leslie Pack, Michael L. Littman, and Andrew W. Moore. "Reinforcement learning: A survey." *Journal of artificial intelligence research* 4 (1996): 237-285.
[157] Barto, Andrew Gehret, Richard S. Sutton, and C. J. C. H. Watkins. *Learning and sequential decision making*. Vol. 89. Amherst, MA: University of Massachusetts, 1989.
[158] Watkins, Christopher JCH, and Peter Dayan. "Q-learning." *Machine learning* 8.3 (1992): 279-292.
[159] Zhinin-Vera, Luis, et al. "Theory of mind and continual reinforcement learning for bullying intervention." *Neural Computing and Applications* 38.6 (2026): 160.
[160] Gallagher, Shaun. "The practice of mind. Theory, simulation or primary interaction?." *Journal of consciousness studies* 8.5-6 (2001): 83-108.
[161] Shanton, Karen, and Alvin Goldman. "Simulation theory." *Wiley Interdisciplinary Reviews: Cognitive Science* 1.4 (2010): 527-538.
[162] Gallese, Vittorio, and Alvin Goldman. "Mirror neurons and the simulation theory of mind-reading." *Trends in cognitive sciences* 2.12 (1998): 493-501.
[163] Harbers, Maaike, Karel Van den Bosch, and John-Jules Meyer. "Modeling agents with a theory of mind: Theory–theory versus simulation theory." *Web Intelligence and Agent Systems* 10.3 (2012): 331-343.
[164] Wellman, Henry M. "Theory of mind: The state of the art." *European Journal of Developmental Psychology* 15.6 (2018): 728-755.
[165] Fisch, Gene S. "Autism and epistemology IV: Does autism need a theory of mind?." *American Journal of Medical Genetics Part A* 161.10 (2013): 2464-2480.
[166] Ilgaz, Hande, and Jedediah WP Allen. "(Co-) Constructing a theory of mind: From language or through language?." *Synthese* 198.9 (2021): 8463-8484.
[167] Wang, Zhenlin, and Lamei Wang. "Cognitive development: Child education." *International encyclopedia of the social & behavioral sciences* 4.2 (2015): 38-42.

[168] Nguyen, Vu-Linh, Mohammad Hossein Shaker, and Eyke Hüllermeier. "How to measure uncertainty in uncertainty sampling for active learning." *Machine Learning* 111.1 (2022): 89-122.
[169] Liddle, Bethany, and Daniel Nettle. "Higher-order theory of mind and social competence in school-age children." *Journal of Cultural and Evolutionary Psychology* 4.3-4 (2006): 231-244.
[170] Wagner, Eitan, et al. "Mind your theory: Theory of mind goes deeper than reasoning." *Findings of the Association for Computational Linguistics: ACL 2025*. 2025.
[171] PERNER, J., and WIMMER, H. (1985): John Thinks That Mary Thinks That - Attribution of 2ndOrder Beliefs by 5-Year-Old to 10-Year-Old Children. Journal of Experimental Child Psychology, 39, 437–471.
[172] De Weerd, Harmen, Rineke Verbrugge, and Bart Verheij. "Higher-order theory of mind is especially useful in unpredictable negotiations." *Autonomous Agents and Multi-Agent Systems* 36.2 (2022): 30.
[173] De Weerd, Harmen, Rineke Verbrugge, and Bart Verheij. "Higher-order theory of mind in negotiations under incomplete information." *International Conference on Principles and Practice of Multi-Agent Systems*. Berlin, Heidelberg: Springer Berlin Heidelberg, 2013.
[174] De Weerd, Harmen, and Bart Verheij. "The advantage of higher-order theory of mind in the game of limited bidding." *Proceedings Workshop 'Reasoning about other Minds', CEUR Workshop Proceedings*. Vol. 751. 2011.
[175] Kılıçel, Senay, and Oğuzhan Kılıçel. "Theory of mind deficit in adolescents with major depressive disorder." (2019).
[176] Wang, Kaizheng, et al. "A Review of Uncertainty Representation and Quantification in Neural Networks." *IEEE Transactions on Pattern Analysis and Machine Intelligence* (2025).
[177] Shafer, Glenn, and Vladimir Vovk. "A tutorial on conformal prediction." *Journal of machine learning research* 9.3 (2008).
[178] Ji, Yuwang, et al. "A survey on tensor techniques and applications in machine learning." *IEEE Access* 7 (2019): 162950-162990.
[179] Asadi, Kavosh, and Michael L. Littman. "An alternative softmax operator for reinforcement learning." *International Conference on Machine Learning*. PMLR, 2017.
[180] Wang, Kaizheng, et al. "Credal deep ensembles for uncertainty quantification." *Advances in Neural Information Processing Systems* 37 (2024): 79540-79572.
[181] Wang, Kaizheng, et al. "Credal ensemble distillation for uncertainty quantification." *Proceedings of the AAAI Conference on Artificial Intelligence*. Vol. 40. No. 31. 2026.
[182] Wang, Kaizheng, et al. "Creinns: Credal-set interval neural networks for uncertainty estimation in classification tasks." *Neural networks* 185 (2025): 107198.
[183] Kudukkil Manchingal, Shireen, et al. "Random-set neural networks." *International Conference on Learning Representations*. Vol. 2025. 2025.
[184] Manchingal, Shireen Kudukkil; Mubashar, Muhammad; Wang, Kaizheng; Shariatmadar, Keivan; Cuzzolin, Fabio. "Random-set convolutional neural network (rs-cnn) for epistemic deep learning." (2023).
[185] Chisholm, James S. "Uncertainty, contingency, and attachment: A life history theory of theory of mind." *From Mating to Mentality*. Psychology Press, 2004. 125-153.
[186] Sarkadi, Ştefan, et al. "Towards an approach for modelling uncertain theory of mind in multi-agent systems." *International Conference on Agreement Technologies*. Cham: Springer International Publishing, 2018.
[187] Bohl, Vivian, and Nivedita Gangopadhyay. "Theory of mind and the unobservability of other minds." *Philosophical Explorations* 17.2 (2014): 203-222.
[188] Hobson, R. Peter. "Against the theory of 'theory of mind'." *British Journal of Developmental Psychology* 9.1 (1991): 33-51.
[189] De Weerd, Harmen, Rineke Verbrugge, and Bart Verheij. "Higher-order theory of mind is especially useful in unpredictable negotiations." *Autonomous Agents and Multi-Agent Systems* 36.2 (2022): 30.
[190] Cuzzolin, Fabio. "Uncertainty measures: A critical survey." *Information Fusion* 114 (2025): 102609.
[191] Hüllermeier, Eyke, and Willem Waegeman. "Aleatoric and epistemic uncertainty in machine learning: An introduction to concepts and methods." *Machine learning* 110.3 (2021): 457-506.
[192] Cuzzolin, Fabio, and Maryam Sultana. *Epistemic Uncertainty in Artificial Intelligence*. Springer Nature Switzerland, 2024.
[193] Sale, Yusuf, Michele Caprio, and Eyke Höllermeier. "Is the volume of a credal set a good measure for epistemic uncertainty?." *Uncertainty in Artificial Intelligence*. PMLR, 2023.
[194] Cuzzolin, Fabio. *The geometry of uncertainty: the geometry of imprecise probabilities*. Springer Nature, 2020.
[195] Jiménez, Sebastián, Mira Jürgens, and Willem Waegeman. "Why machine learning models fail to fully capture epistemic uncertainty." *arXiv preprint arXiv:2505.23506* (2025).
[196] Matta, Daoud. "Artificial intelligence and theory of mind." *Journal of Psychology and AI* 2.1 (2026): 2628373.
[197] Fernandes, Shannon J., and Tejeshwar Dhananjaya. "Prediction Machines–Comparative Analysis of Theory of Mind Abilities in Machine Learning Models." *International Journal of Indian Psychȯlogy* 11.3 (2023).
[198] Lică, Mircea, et al. "Mindforge: Empowering embodied agents with theory of mind for lifelong cultural learning." *Advances in Neural Information Processing Systems* 38 (2026): 170024-170053.
[199] Wang, Liyuan, et al. "A comprehensive survey of continual learning: Theory, method and application." *IEEE transactions on pattern analysis and machine intelligence* 46.8 (2024): 5362-5383.
[200] Zhinin-Vera, Luis, et al. "Theory of mind and continual reinforcement learning for bullying intervention." *Neural Computing and Applications* 38.6 (2026): 160.
[201] Lee, Soochan, et al. "Learning to continually learn with the Bayesian principle." *arXiv preprint arXiv:2405.18758* (2024).
[202] Adel, Tameem. "The bayesian approach to continual learning: An overview." *arXiv preprint arXiv:2507.08922* (2025).

[203] Kuhn, Deanna. "Theory of mind, metacognition, and reasoning: A life-span perspective." *Children's reasoning and the mind*. Psychology Press, 2014. 301-326.
[204] Iordanou, Kalypso. "From theory of mind to epistemic cognition. A lifespan perspective." *Frontline Learning Research* 4.5 (2016): 106-119.
[205] Knutsen, John, Douglas Frye, and David M. Sobel. "Theory of learning, theory of teaching, and theory of mind: How social-cognitive development influences children's understanding of learning and teaching." (2013).
[206] Giovagnoli, Anna Rita. "Theory of mind across lifespan from ages 16 to 81 years." *Epilepsy & Behavior* 100 (2019): 106349.
[207] Langley, Christelle, et al. "Theory of mind and preference learning at the interface of cognitive science, neuroscience, and AI: A review." *Frontiers in artificial intelligence* 5 (2022): 778852.
[208] Zeng, Yi, et al. "A brain-inspired model of theory of mind." *Frontiers in Neurorobotics* 14 (2020): 60.
[209] Kudithipudi, Dhireesha, et al. "Biological underpinnings for lifelong learning machines." *Nature Machine Intelligence* 4.3 (2022): 196-210.
[210] Zhao, Changyuan, et al. "From agentification to self-evolving agentic AI for wireless networks: Concepts, approaches, and future research directions." *IEEE Communications Magazine* (2026).
[211] Caroline, M. Victoriya, and G. Maria Priscilla. "Agentic Artificial Intelligence: Frameworks for Autonomous Decision-Making and Continuous Self-Improvement."
[212] Bar, Kaushik. "Self-improving agentic AI through Bayesian meta-learning." *Available at SSRN 5190135* (2025).
[213] G. Zhang et al., "Evoflow: Evolving Diverse Agentic Workflows on the Fly," arXiv preprint arXiv:2502.07373, 2025.
[214] H. Sun et al., "Adaplanner: Adaptive Planning From Feedback with Language Models," Advances in Neural Information Processing Systems, vol. 36, 2023, pp. 58,202–45.
[215] J. Fang et al., "A Comprehensive Survey of Self-Evolving AI Agents: A New Paradigm Bridging Foundation Models and Lifelong Agentic Systems," arXiv preprint arXiv:2508.07407, 2025.
[216] H.-a. Gao et al., "A Survey of Self-Evolving Agents: On Path to Artificial Super Intelligence," arXiv preprint arXiv:2507.21046, 2025.
[217] Keysers, Christian, and Valeria Gazzola. "Integrating simulation and theory of mind: from self to social cognition." *Trends in cognitive sciences* 11.5 (2007): 194-196.
[218] Perrykkad, Kelsey, and Jakob Hohwy. "Modelling me, modelling you: the autistic self." *Review Journal of Autism and Developmental Disorders* 7.1 (2020): 1-31.
[219] Vogeley, Kai, et al. "Mind reading: neural mechanisms of theory of mind and self-perspective." *Neuroimage* 14.1 (2001): 170-181.
[220] Moutoussis, Michael, et al. "Bayesian inferences about the self (and others): A review." *Consciousness and cognition* 25 (2014): 67-76.
[221] Zeng, Yi, et al. "A brain-inspired model of theory of mind." *Frontiers in Neurorobotics* 14 (2020): 60.
[222] Winfield, Alan FT. "Experiments in artificial theory of mind: From safety to story-telling." *Frontiers in Robotics and AI* 5 (2018): 357467.
[223] Harbers, Maaike, Karel Van den Bosch, and John-Jules Meyer. "Modeling agents with a theory of mind: Theory–theory versus simulation theory." *Web Intelligence and Agent Systems* 10.3 (2012): 331-343.
[224] Li, Y., Wei, F., Zhao, J., Zhang, C., and Zhang, H. Rain: Your language models can align themselves without finetuning. In The Twelfth International Conference on Learning Representations, 2024.
[225] Sun, Z., Shen, Y., Zhou, Q., Zhang, H., Chen, Z., Cox, D. D., Yang, Y., and Gan, C. Principle-driven self-alignment of language models from scratch with minimal human supervision. In Thirty-seventh Conference on Neural Information Processing Systems, 2023.
[226] Berglund, L., Stickland, A. C., Balesni, M., Kaufmann, M., Tong, M., Korbak, T., Kokotajlo, D., and Evans, O. Taken out of context: On measuring situational awareness in llms. arXiv preprint arXiv:2309.00667, 2023.
[227] Chen, Dongping, et al. "Self-cognition in large language models: An exploratory study." *arXiv preprint arXiv:2407.01505* (2024).
[228] Pynadath, David V., Paul S. Rosenbloom, and Stacy C. Marsella. "Reinforcement learning for adaptive theory of mind in the sigma cognitive architecture." *International Conference on Artificial General Intelligence*. Cham: Springer International Publishing, 2014.
[229] Oguntola, Ini, et al. "Theory of mind as intrinsic motivation for multi-agent reinforcement learning." *arXiv preprint arXiv:2307.01158* (2023).
[230] Nguyen, Thuy Ngoc, and Cleotilde Gonzalez. "Theory of mind from observation in cognitive models and humans." *Topics in cognitive science* 14.4 (2022): 665-686.
[231] Westby, Samuel, and Christoph Riedl. "Collective intelligence in human-AI teams: A Bayesian theory of mind approach." *Proceedings of the AAAI Conference on Artificial Intelligence*. Vol. 37. No. 5. 2023.
[232] Kleiman-Weiner, M., Ho, M. K., Austerweil, J. L., Littman, M. L., & Tenenbaum, J. B. (2016). Coordinate to cooperate or compete: Abstract
goals and joint intentions in social interaction. In D. Mirman, A. Papafragou, D. Grodner, & J. C. Trueswell (Eds.), Proceedings of the
38th Annual Conference of the Cognitive Science Society (pp. 1679–1684). Cognitive Science Society.

[233] Ying, Lance, et al. "Understanding epistemic language with a language-augmented bayesian theory of mind." *Transactions of the Association for Computational Linguistics* 13 (2025): 613-637.
[234] Cheng, Yuheng, et al. "Exploring large language model based intelligent agents: Definitions, methods, and prospects." *arXiv preprint arXiv:2401.03428* (2024).
[235] Guo, Taicheng, et al. "Large language model based multi-agents: A survey of progress and challenges." *arXiv preprint arXiv:2402.01680* (2024).
[236] Xi, Zhiheng, et al. "Agentgym: Evaluating and training large language model-based agents across diverse environments." *Proceedings of the 63rd Annual Meeting of the Association for Computational Linguistics (Volume 1: Long Papers)*. 2025.
[237] Sarangi, Sneheel, et al. "Agentic-ToM: Cognition-Inspired Agentic Processing For Enhancing Theory of Mind Reasoning in Large Language Models." *30th Conference on Empirical Methods in Natural Language Processing, EMNLP 2025*. Association for Computational Linguistics (ACL), 2025.
[238] Harbers, Maaike, Karel Van den Bosch, and John-Jules Meyer. "Modeling agents with a theory of mind: Theory–theory versus simulation theory." *Web Intelligence and Agent Systems* 10.3 (2012): 331-343.
[239] Gurney, Nikolos, et al. "Operationalizing theories of theory of mind: A survey." *AAAI fall symposium*. Cham: Springer Nature Switzerland, 2021.
[240] Rocha, Michele, et al. "Applying theory of mind to multi-agent systems: A systematic review." *Brazilian Conference on Intelligent Systems*. Springer, Cham, 2023.
[241] Zhinin-Vera, Luis, et al. "Mindful human digital twins: integrating theory of mind with multi-agent reinforcement learning." *Applied Soft Computing* 174 (2025): 112939.
[242] Winfield, Alan FT. "Experiments in artificial theory of mind: From safety to story-telling." *Frontiers in Robotics and AI* 5 (2018): 357467.
[243] Harbers, Maaike, Karel Van den Bosch, and John-Jules Meyer. "Modeling agents with a theory of mind: Theory–theory versus simulation theory." *Web Intelligence and Agent Systems* 10.3 (2012): 331-343.
[244] Katt, Sammie, and Samuel Kaski. "Artificial Theory of Mind in Human-in-the-Loop." *EurIPS 2025 Workshop on Metacognition in Generative AI*.
[245] Keurulainen, Oskar, Gokhan Alcan, and Ville Kyrki. "The role of higher-order cognitive models in active learning." *arXiv preprint arXiv:2401.04397* (2024).
[246] Ghiglino, Davide, et al. "Enhancing theory of mind in autism through humanoid robot interaction in a randomized controlled trial." *Scientific Reports* 15.1 (2025): 27650.
[247] Yang, Scott Cheng-Hsin, et al. "A unifying computational framework for teaching and active learning." *Topics in cognitive science* 11.2 (2019): 316-337.
[248] Abdelghani, Rania, Hélène Sauzéon, and Pierre-Yves Oudeyer. "Generative AI in the classroom: Can students remain active learners?." *arXiv preprint arXiv:2310.03192* (2023).
[249] Ghai, Bhavya, et al. "Explainable active learning (xal) toward ai explanations as interfaces for machine teachers." *Proceedings of the ACM on human-computer interaction* 4.CSCW3 (2021): 1-28.
[250] Kim, Kyung-Joong, and Hod Lipson. "Towards a" theory of mind" in simulated robots." *Proceedings of the 11th Annual Conference Companion on Genetic and Evolutionary Computation Conference: Late Breaking Papers*. 2009.
[251] Yufan Wu, Yinghui He, Yilin Jia, Rada Mihalcea, Yulong Chen, and Naihao Deng. 2023. Hi-ToM: A benchmark for evaluating higher-order theory of mind reasoning in large language models. In Findings of the Association for Computational Linguistics: EMNLP 2023, pages 10691–10706, Singapore. Association for Computational Linguistics.
[252] Piotr J Gmytrasiewicz and Prashant Doshi. 2005. A framework for sequential planning in multi-agent settings. Journal of Artificial Intelligence Research, 24:49–79.
[253] Yoshida, Wako, Ray J. Dolan, and Karl J. Friston. "Game theory of mind." *PLoS computational biology* 4.12 (2008): e1000254.
[254] Harré, Michael S. "What can game theory tell us about an AI 'theory of mind'?." *Games* 13.3 (2022): 46.
[255] de Weerd, Harmen, Rineke Verbrugge, and Bart Verheij. "Agent-based models for higher-order theory of mind." *Advances in Social Simulation: Proceedings of the 9th Conference of the European Social Simulation Association*. Berlin, Heidelberg: Springer Berlin Heidelberg, 2014.
[256] Hongxin Zhang, Weihua Du, Jiaming Shan, Qinhong Zhou, Yilun Du, Joshua B Tenenbaum, Tianmin Shu, and Chuang Gan. 2023. Building cooperative embodied agents modularly with large language models. arXiv preprint arXiv:2307.02485.
[257] utility-proportional beliefs (UPB) agents
[258] Wu, Yufan, et al. "Hi-tom: A benchmark for evaluating higher-order theory of mind reasoning in large language models." *Findings of the Association for Computational Linguistics: EMNLP 2023*. 2023.
[259] Kosinski, Michal. "Evaluating large language models in theory of mind tasks." *Proceedings of the National Academy of Sciences* 121.45 (2024): e2405460121.
[260] Song, Li, Kun Yang, and Chao Chen. "Adversarial Distributed Multi-Task Meta-Inverse Reinforcement Learning with Theory of Mind and Mean-Field Method." *Mathematics* 14.4 (2026): 691.
[261] Yu, Chuang, Baris Serhan, and Angelo Cangelosi. "ToP-ToM: Trust-aware robot policy with theory of mind." *2024 IEEE International Conference on Robotics and Automation (ICRA)*. IEEE, 2024.
[262] Westby, Samuel, and Christoph Riedl. "Collective intelligence in human-AI teams: A Bayesian theory of mind approach." *Proceedings of the AAAI Conference on Artificial Intelligence*. Vol. 37. No. 5. 2023.

[263] Saxe, Rebecca, and Sean Dae Houlihan. "Formalizing emotion concepts within a Bayesian model of theory of mind." *Current opinion in Psychology* 17 (2017): 15-21.
[264] Baker, Chris, Rebecca Saxe, and Joshua Tenenbaum. "Bayesian theory of mind: Modeling joint belief-desire attribution." *Proceedings of the annual meeting of the cognitive science society*. Vol. 33. No. 33. 2011.
[265] Spaan, Matthijs TJ. "Partially observable Markov decision processes." *Reinforcement learning: State-of-the-art*. Berlin, Heidelberg: Springer Berlin Heidelberg, 2012. 387-414.
[266] Cho, Isu, Nellie Kamkar, and Niki Hosseini-Kamkar. "Reasoning about mental states under uncertainty." *PloS one* 17.11 (2022): e0277356.
[267] Sarkadi, Ştefan, et al. "Towards an approach for modelling uncertain theory of mind in multi-agent systems." *International Conference on Agreement Technologies*. Cham: Springer International Publishing, 2018.
[268] Sicilia, Anthony, and Malihe Alikhani. "Evaluating Theory of (an uncertain) Mind: Predicting the Uncertain Beliefs of Others in Conversation Forecasting." *arXiv preprint arXiv:2409.14986* (2024).
[269] Sicilia, Anthony, et al. "Deal, or no deal (or who knows)? forecasting uncertainty in conversations using large language models." *Findings of the Association for Computational Linguistics: ACL 2024*. 2024.
[270] Manchingal, Shireen Kudukkil, and Fabio Cuzzolin. "Epistemic deep learning." *arXiv preprint arXiv:2206.07609* (2022).
[271] Manchingal, Shireen Kudukkil, et al. "Epistemic Artificial Intelligence is Essential for Machine Learning Models to Truly'Know When They Do Not Know'." *arXiv preprint arXiv:2505.04950* (2025).
[272] Manchingal, Shireen Kudukkil, et al. "A unified evaluation framework for epistemic predictions." *arXiv preprint arXiv:2501.16912* (2025).
[273] Kudukkil Manchingal, Shireen, et al. "Random-set neural networks." *International Conference on Learning Representations*. Vol. 2025. 2025.
[274] Wang, Kaizheng, et al. "Creinns: Credal-set interval neural networks for uncertainty estimation in classification tasks." *Neural networks* 185 (2025): 107198.
[275] Wang, Kaizheng, et al. "Credal ensemble distillation for uncertainty quantification." *Proceedings of the AAAI Conference on Artificial Intelligence*. Vol. 40. No. 31. 2026.
[276] Caprio, Michele, et al. "Credal learning theory." *Advances in Neural Information Processing Systems* 37 (2024): 38665-38694.
[277] Pynadath, David V., and Stacy C. Marsella. "PsychSim: Modeling theory of mind with decision-theoretic agents." *IJCAI*. Vol. 5. 2005.
[278] Yoshida, Wako, Ray J. Dolan, and Karl J. Friston. "Game theory of mind." *PLoS computational biology* 4.12 (2008): e1000254.
[279] Rocha, Michele, et al. "Applying theory of mind to multi-agent systems: A systematic review." *Brazilian Conference on Intelligent Systems*. Springer, Cham, 2023.
[280] Panisson, Alison R., et al. "On the formal semantics of theory of mind in agent communication." *International Conference on Agreement Technologies*. Cham: Springer International Publishing, 2018.
[281] Byom, Lindsey J., and Bilge Mutlu. "Theory of mind: Mechanisms, methods, and new directions." *Frontiers in human neuroscience* 7 (2013): 413.
[282] Cross, Logan, et al. "Hypothetical minds: Scaffolding theory of mind for multi-agent tasks with large language models." *International Conference on Learning Representations*. Vol. 2025. 2025.
[283] Apperly, Ian A. "What is "theory of mind"? Concepts, cognitive processes and individual differences." *Quarterly journal of experimental psychology* 65.5 (2012): 825-839.
[284] Dunbar, Robin IM. "Theory of mind and the evolution of language." *Approaches to the Evolution of Language* (1998): 92-110.
[285] Lim, Terence X., Sidney Tio, and Desmond C. Ong. "Improving multi-agent cooperation using theory of mind." *arXiv preprint arXiv:2007.15703* (2020).
[286] Mason, M. F., Cloutier, J., and Macrae, C. N. (2006). On construing others: Category and stereotype 1014 activation from facial cues. *Social Cognition*, 24(5), 540-562.
[287] F. Cuzzolin and A. Morelli. "From Psychology and Neuroscience to AI: Toward a Learning-Based Computational Theory of Mind Model". Preprint 26 May 2026. DOI:10.20944/preprints202605.1676.v1
[288] Zwickel, Jan. "Agency attribution and visuospatial perspective taking." *Psychonomic Bulletin & Review* 16.6 (2009): 1089-1093.
[289] Taylor, Derry, et al. "Reading minds or reading scripts? De-intellectualising theory of mind." *Biological Reviews* 98.6 (2023): 2028-2048.
[290] Johnson, Susan C. "Detecting agents." *Philosophical Transactions of the Royal Society of London. Series B: Biological Sciences* 358.1431 (2003): 549-559.
[291] Bourgine, Paul. "What is cognitive economics?." *Cognitive economics: An interdisciplinary approach*. Berlin, Heidelberg: Springer Berlin Heidelberg, 2004. 1-12.
[292] Ren, Pengzhen, et al. "A comprehensive survey of neural architecture search: Challenges and solutions." *ACM Computing Surveys (CSUR)* 54.4 (2021): 1-34.
[293] Liu, Yuqiao, et al. "A survey on evolutionary neural architecture search." *IEEE transactions on neural networks and learning systems* 34.2 (2021): 550-570.
[294] Liu, Chenxi, et al. "Progressive neural architecture search." *Proceedings of the European conference on computer vision (ECCV)*. 2018.

[295] Liu, Yuqiao, et al. "A survey on evolutionary neural architecture search." *IEEE transactions on neural networks and learning systems* 34.2 (2021): 550-570.
[296] Fan, Yicheng, et al. "Layernas: Neural architecture search in polynomial complexity." *arXiv preprint arXiv:2304.11517* (2023).
[297] Jin, Haifeng, Qingquan Song, and Xia Hu. "Auto-keras: An efficient neural architecture search system." *Proceedings of the 25th ACM SIGKDD international conference on knowledge discovery & data mining*. 2019.
[298] Rocha, Michele, et al. "Applying theory of mind to multi-agent systems: A systematic review." *Brazilian Conference on Intelligent Systems*. Springer, Cham, 2023.
[299] Matta, Daoud. "Artificial intelligence and theory of mind." *Journal of Psychology and AI* 2.1 (2026): 2628373.
[300] Shalev-Shwartz, Shai. "Online learning and online convex optimization." *Foundations and Trends® in Machine Learning* 4.2 (2025): 107-194.
[301] Hayes, Conor F., et al. "A practical guide to multi-objective reinforcement learning and planning: CF Hayes et al." *Autonomous Agents and Multi-Agent Systems* 36.1 (2022): 26.
[302] Borsci, Simone, et al. "The chatbot usability scale: the design and pilot of a usability scale for interaction with AI-based conversational agents." *Personal and ubiquitous computing* 26.1 (2022): 95-119.
[303] Marconi, Luca, Luca Longo, and Federico Cabitza. "Assessing Interaction Quality in Human–AI Dialogue: An Integrative Review and Multi-Layer Framework for Conversational Agents." *Machine Learning and Knowledge Extraction* 8.2 (2026): 28.
[304] Nevin, Rebecca, Aleksandra Ćiprijanović, and Brian D. Nord. "Deepuq: Assessing the aleatoric uncertainties from two deep learning methods." *arXiv preprint arXiv:2411.08587* (2024).
[305] Juergens, Mira, et al. "Is epistemic uncertainty faithfully represented by evidential deep learning methods?." *arXiv preprint arXiv:2402.09056* (2024).
[306] Mubashar, Muhammad, and Fabio Cuzzolin. "Epistemic Generative Adversarial Networks." *arXiv preprint arXiv:2603.18348* (2026).
[307] Chakraborty, Tanujit, et al. "Ten years of generative adversarial nets (GANs): a survey of the state-of-the-art." *Machine Learning: Science and Technology* 5.1 (2024): 011001.
[308] Doersch, Carl. "Tutorial on variational autoencoders." *arXiv preprint arXiv:1606.05908* (2016).
[309309] Andreas, Jacob, et al. "Neural module networks." *Proceedings of the IEEE conference on computer vision and pattern recognition*. 2016.
[310] Yao, Lewei, et al. "SM-NAS: Structural-to-modular neural architecture search for object detection." *Proceedings of the AAAI conference on artificial intelligence*. Vol. 34. No. 07. 2020.
[311] Chen, Yaran, et al. "ModuleNet: Knowledge-inherited neural architecture search." *IEEE Transactions on Cybernetics* 52.11 (2021): 11661-11671.
[312] Harbers, Maaike, Karel Van Den Bosch, and John-Jules Meyer. "Modeling agents with a theory of mind." *2009 IEEE/WIC/ACM International Joint Conference on Web Intelligence and Intelligent Agent Technology*. Vol. 2. IEEE, 2009.
[313] Xue, Chao, et al. "On neural architecture search and hyperparameter optimization: A max-flow based approach." *Neural Networks* 188 (2025): 107507.
[314] Zoph, Barret; Le, Quoc V. (2016-11-04). "Neural Architecture Search with Reinforcement Learning". arXiv:1611.01578
[315] Zoph, Barret; Vasudevan, Vijay; Shlens, Jonathon; Le, Quoc V. (2017-07-21). "Learning Transferable Architectures for Scalable Image Recognition". arXiv:1707.07012
[316] Gao, Qiang, et al. "Efficient architecture search for continual learning." *IEEE Transactions on Neural Networks and Learning Systems* 34.11 (2022): 8555-8565.
[317] Huang, Shenyang, Vincent François-Lavet, and Guillaume Rabusseau. "Neural architecture search for class-incremental learning." *arXiv preprint arXiv:1909.06686* (2019).
[318] Shahawy, Mohamed, Elhadj Benkhelifa, and David White. "Exploring the intersection between neural architecture search and continual learning." *IEEE Transactions on Neural Networks and Learning Systems* 36.7 (2024): 11776-11792.
[319] Chen, Yaran, et al. "ModuleNet: Knowledge-inherited neural architecture search." *IEEE Transactions on Cybernetics* 52.11 (2021): 11661-11671.
[320] Gurney, Nikolos, et al. "Operationalizing theories of theory of mind: A survey." *AAAI fall symposium*. Cham: Springer Nature Switzerland, 2021.
[321] Schaul, Tom; Schmidhuber, Jürgen (2010). "Metalearning". *Scholarpedia*. **5** (6): 4650
[322] Zhou, Fengwei, Bin Wu, and Zhenguo Li. "Deep meta-learning: Learning to learn in the concept space." *arXiv preprint arXiv:1802.03596* (2018).
[323] Utgoff, Paul E. *Machine learning of inductive bias*. Springer Science & Business Media, 2012.
[324] He, Xin, Kaiyong Zhao, and Xiaowen Chu. "AutoML: A survey of the state-of-the-art." *Knowledge-based systems* 212 (2021): 106622.
[325] Barbudo, Rafael, Sebastián Ventura, and José Raúl Romero. "Eight years of AutoML: categorisation, review and trends." *Knowledge and Information Systems* 65.12 (2023): 5097-5149.
[326] Hazan, Elad. "Introduction to online convex optimization." *Foundations and Trends in Optimization* 2.3-4 (2016): 157-325.
[327] Shalev-Shwartz, Shai. "Online learning and online convex optimization." *Foundations and Trends® in Machine Learning* 4.2 (2025): 107-194.
[328] Shahawy, Mohamed, Elhadj Benkhelifa, and David White. "Exploring the intersection between neural architecture search and continual learning." *IEEE Transactions on Neural Networks and Learning Systems* 36.7 (2024): 11776-11792.

[329] Siegal, Michael, and Rosemary Varley. "Neural systems involved in'theory of mind'." *Nature Reviews Neuroscience* 3.6 (2002): 463-471.
[330] Harrison, David. *Mind the Matter: Active Matter, Basal Cognition, and the Making of Bio-Inspired Artificial Intelligence*. Diss. 2024.
[331] Zeng, Yi, et al. "A brain-inspired model of theory of mind." *Frontiers in Neurorobotics* 14 (2020): 60.
[332] Nebreda, Alberto, et al. "The social machine: artificial intelligence (AI) approaches to theory of mind." *The theory of mind under scrutiny: psychopathology, neuroscience, philosophy of mind and artificial intelligence*. Cham: Springer Nature Switzerland, 2024. 681-722.
[333] Gallese, Vittorio. "Before and below 'theory of mind': embodied simulation and the neural correlates of social cognition." *Philosophical Transactions of the Royal Society B: Biological Sciences* 362.1480 (2007): 659-669.
[334] Zhao, Zhuoya, et al. "A brain-inspired theory of mind spiking neural network improves multi-agent cooperation and competition." *Patterns* 4.8 (2023).
[335] Gorgan Mohammadi, Ashena, and Mohammad Ganjtabesh. "On computational models of theory of mind and the imitative reinforcement learning in spiking neural networks." *Scientific Reports* 14.1 (2024): 1945.
[336] Borbás, Réka, et al. "Neural correlates of theory of mind in children and adults using CAToon: Introducing an open-source child-friendly neuroimaging task." *Developmental Cognitive Neuroscience* 49 (2021): 100959.
[337] Schaafsma, Sara M., et al. "Deconstructing and reconstructing theory of mind." *Trends in cognitive sciences* 19.2 (2015): 65-72.
[338] Siegal, Michael, and Rosemary Varley. "Neural systems involved in'theory of mind'." *Nature Reviews Neuroscience* 3.6 (2002): 463-471.
[339] Deco, Gustavo, and Morten L. Kringelbach. *Whole-brain modelling: Cartography of the dynamics of mind*. Oxford University Press, 2026.
[340] Van Der Molen, Tjitse, et al. "Preconfigured neuronal firing sequences in human brain organoids." *Nature neuroscience* 29.1 (2026): 123-135.
[341] Li, Jiahe, et al. "Assembling the Mind's Mosaic: Towards EEG Semantic Intent Decoding." *arXiv preprint arXiv:2601.20447* (2026).
[342] Rouder, Jeffrey N., Richard D. Morey, and Michael S. Pratte. "Bayesian hierarchical models." *New handbook of mathematical psychology* 1 (2014).
[343] Simons, Kenneth W. "Rethinking mental states." *BUL Rev.* 72 (1992): 463.
[344] Rouder, Jeffrey N., Richard D. Morey, and Michael S. Pratte. "Bayesian hierarchical models." *New handbook of mathematical psychology* 1 (2014).
[345] Williams, Daniel. "Hierarchical Bayesian models of delusion." *Consciousness and cognition* 61 (2018): 129-147.
[346] Zhou, Jie, et al. "Graph neural networks: A review of methods and applications." *AI open* 1 (2020): 57-81.
[347] Xu, Keyulu, et al. "How powerful are graph neural networks?." *arXiv preprint arXiv:1810.00826* (2018).
[348] Sun, Xiangguo, et al. "All in one: Multi-task prompting for graph neural networks." *Proceedings of the 29th ACM SIGKDD conference on knowledge discovery and data mining*. 2023.
[349] Zhang, Chuxu, et al. "Heterogeneous graph neural network." *Proceedings of the 25th ACM SIGKDD international conference on knowledge discovery & data mining*. 2019.
[350] Woodley, Tommy, et al. "Random-Set Graph Neural Networks." *arXiv preprint arXiv:2605.11987* (2026).
[351] Krause, Laura, et al. "The role of medial prefrontal cortex in theory of mind: a deep rTMS study." *Behavioural brain research* 228.1 (2012): 87-90.
[352] Otti, Alexander, Afra M. Wohlschlaeger, and Michael Noll-Hussong. "Is the medial prefrontal cortex necessary for theory of mind?." *PloS one* 10.8 (2015): e0135912.
[353] Georgeff, Michael, et al. "The belief-desire-intention model of agency." *International workshop on agent theories, architectures, and languages*. Berlin, Heidelberg: Springer Berlin Heidelberg, 1998.
[354] Jafari, Mehdi, et al. "Beyond Words: Integrating Theory of Mind into Conversational Agents for Human-Like Belief, Desire, and Intention Alignment." *Findings of the Association for Computational Linguistics: ACL 2025*. 2025.
[355] Deb, Kalyanmoy, Karthik Sindhya, and Jussi Hakanen. "Multi-objective optimization." *Decision sciences*. CRC Press, 2016. 161-200.
[356] Gunantara, Nyoman. "A review of multi-objective optimization: Methods and its applications." *Cogent Engineering* 5.1 (2018): 1502242.
[357] Rădulescu, Roxana, et al. "Multi-objective multi-agent decision making: a utility-based analysis and survey." *Autonomous Agents and Multi-Agent Systems* 34.1 (2020): 10.
[358] Stanimirovic, Ivan P., Milan Lj Zlatanovic, and Marko D. Petkovic. "On the linear weighted sum method for multi-objective optimization." *Facta Acta Univ* 26.4 (2011): 49-63.
[359] Kalita, Kanak, et al. "Multi-objective exponential distribution optimizer (MOEDO): a novel math-inspired multi-objective algorithm for global optimization and real-world engineering design problems." *Scientific reports* 14.1 (2024): 1816.
[360] Watkins, Christopher JCH, and Peter Dayan. "Q-learning." *Machine learning* 8.3 (1992): 279-292.
[361] Puterman, Martin L. "Markov decision processes." *Handbooks in operations research and management science* 2 (1990): 331-434.

[362] Emmert-Streib, Frank, and Matthias Dehmer. "Understanding statistical hypothesis testing: The logic of statistical inference." *Machine Learning and Knowledge Extraction* 1.3 (2019): 945-962.
[363] Cuzzolin, Fabio. *The geometry of uncertainty: the geometry of imprecise probabilities*. Springer Nature, 2020.
[364] Tootchi, Amirreza, and Xiaoping Du. "Robust design under machine learning model uncertainty." *Journal of Mechanical Design* 148.2 (2026): 021701.
[365] Khetarpal, Khimya, et al. "Towards continual reinforcement learning: A review and perspectives." *Journal of Artificial Intelligence Research* 75 (2022): 1401-1476.
[366] Kessler, Samuel, et al. "Same state, different task: Continual reinforcement learning without interference." *Proceedings of the AAAI Conference on Artificial Intelligence*. Vol. 36. No. 7. 2022.
[367] Hoi, Steven CH, et al. "Online learning: A comprehensive survey." *Neurocomputing* 459 (2021): 249-289.
[368] Hazan, Elad. "Introduction to online convex optimization." *Foundations and Trends in Optimization* 2.3-4 (2016): 157-325.
[369] Wang, Hao, and Dit-Yan Yeung. "A survey on Bayesian deep learning." *ACM computing surveys (csur)* 53.5 (2020): 1-37.
[370] Dong, Xibin, et al. "A survey on ensemble learning." *Frontiers of Computer Science* 14.2 (2020): 241-258.
[371] Angelopoulos, Anastasios N., and Stephen Bates. "Conformal prediction: A gentle introduction." *Foundations and Trends in Machine Learning* 16.4 (2023): 494-591.
[372] Khosravi, Abbas, et al. "Comprehensive review of neural network-based prediction intervals and new advances." *IEEE Transactions on neural networks* 22.9 (2011): 1341-1356.
[373] Sensoy, Murat; Kaplan, Lance; Kandemir, Melih (2018-10-31), *Evidential Deep Learning to Quantify Classification Uncertainty*, arXiv:1806.01768.
[374] Kaizheng Wang, Fabio Cuzzolin, Keivan Shariatmadar, David Moens and Hans Hallez. "A Review of Uncertainty Representation and Quantification in Neural Networks." *IEEE Transactions on Pattern Analysis and Machine Intelligence* (2025).
[375] Mobiny, Aryan; Yuan, Pengyu; Moulik, Supratik K.; Garg, Naveen; Wu, Carol C.; Van Nguyen, Hien (2021-03-09). "DropConnect is effective in modeling uncertainty of Bayesian deep networks". *Scientific Reports*. **11** (1): 5458.
[376] Maohao Shen, J. Jon Ryu, Soumya Ghosh, Yuheng Bu, Prasanna Sattigeri, Subhro Das, Gregory W. Wornell (2024). "Are Uncertainty Quantification Capabilities of Evidential Deep Learning a Mirage?". *Advances in Neural Information Processing Systems* 37.
[377] Datta, Gauri Sankar; Ghosh, Malay (1996-02-01). "On the invariance of noninformative priors". The Annals of Statistics. 24 (1). doi:10.1214/aos/1033066203.
[378] Fragoso, Tiago M., Wesley Bertoli, and Francisco Louzada. "Bayesian model averaging: A systematic review and conceptual classification." *International Statistical Review* 86.1 (2018): 1-28.
[379] Shireen Kudukkil Manchingal, Andrew Bradley, Julian F. P. Kooij, Keivan Shariatmadar, Neil Yorke-Smith, Fabio Cuzzolin. "Epistemic Artificial Intelligence is Essential for Machine Learning Models to Truly 'Know When They Do Not Know'." *arXiv preprint arXiv:2505.04950* (2025).
[380] Shireen Kudukkil Manchingal, Muhammad Mubashar, Kaizheng Wang, Fabio Cuzzolin. "A unified evaluation framework for epistemic predictions." *arXiv preprint arXiv:2501.16912* (2025).
[381] Shireen Kudukkil Manchingal, Muhammad Mubashar, Kaizheng Wang, Keivan Shariatmadar, Fabio Cuzzolin. "Random-set neural networks." *International Conference on Learning Representations*. Vol. 2025. 2025.
[382] Kaizheng Wang, Fabio Cuzzolin, Shireen Kudukkil Manchingal, Keivan Shariatmadar, David Moens, Hans Hallez. "Credal deep ensembles for uncertainty quantification." *Advances in Neural Information Processing Systems* 37 (2024): 79540-79572.
[383] Kaizheng Wang, Fabio Cuzzolin, David Moens, Hans Hallez. "Credal ensemble distillation for uncertainty quantification." *Proceedings of the AAAI Conference on Artificial Intelligence*. Vol. 40. No. 31. 2026.
[384] Kaizheng Wang, Fabio Cuzzolin, Keivan Shariatmadar, David Moens, Hans Hallez. "Credal wrapper of model averaging for uncertainty estimation in classification." *International Conference on Learning Representations*. Vol. 2025. 2025.
[385] Kaizheng Wang, Keivan Shariatmadar, Shireen Kudukkil Manchingal, Fabio Cuzzolin, David Moens and Hans Hallez. "Creinns: Credal-set interval neural networks for uncertainty estimation in classification tasks." *Neural networks* 185 (2025): 107198.
[386] Caprio, Michele, Sultana, Maryam, Elia, Eleni and Cuzzolin, Fabio. "Credal learning theory." *Advances in Neural Information Processing Systems* 37 (2024): 38665-38694.
[387] Osman, Magda. "Controlling uncertainty: a review of human behavior in complex dynamic environments." *Psychological bulletin* 136.1 (2010): 65.
[388] Kurniawati, Hanna, and Vinay Yadav. "An online POMDP solver for uncertainty planning in dynamic environment." *Robotics Research: The 16th International Symposium ISRR*. Cham: Springer International Publishing, 2016.
[389] Cuzzolin, Fabio, and Shireen Kudukkil Manchingal. "A Research Programme for Continual and Neurosymbolic Learning in Epistemic Artificial Intelligence." (2026).
[390] Scriven, Michael. "An essential unpredictability in human behavior." US Information Agency, Voice of America Forum, 1964.
[391] Cziko, Gary A. "Unpredictability and indeterminism in human behavior: Arguments and implications for educational research." *Educational researcher* 18.3 (1989): 17-25.
[392] Herry, Cyril, et al. "Processing of temporal unpredictability in human and animal amygdala." *Journal of Neuroscience* 27.22 (2007): 5958-5966.
[393] Shiffrin, Richard M., et al. "A survey of model evaluation approaches with a tutorial on hierarchical Bayesian methods." *Cognitive Science* 32.8 (2008): 1248-1284.

[394] Georgeff, Michael, et al. "The belief-desire-intention model of agency." *International workshop on agent theories, architectures, and languages*. Berlin, Heidelberg: Springer Berlin Heidelberg, 1998.
[395] Qu, Meng, and Jian Tang. "Probabilistic logic neural networks for reasoning." *Advances in neural information processing systems* 32 (2019).
[396] Lang, Jérôme, Leendert Van Der Torre, and Emil Weydert. "Hidden uncertainty in the logical representation of desires." *IJCAI*. 2003.
[397] Manhaeve, Robin, et al. "Deepproblog: Neural probabilistic logic programming." *Advances in neural information processing systems* 31 (2018).
[398] Tolloso, Matteo, and Davide Bacciu. "Credal Graph Neural Networks." *arXiv preprint arXiv:2512.02722* (2025).
[399] Woodley, Tommy, et al. "Random-Set Graph Neural Networks." *arXiv preprint arXiv:2605.11987* (2026).
[400] Zhang, Yuyu, et al. "Efficient probabilistic logic reasoning with graph neural networks." *arXiv preprint arXiv:2001.11850* (2020).
[401] Shenoy, Prakash P., and Glenn Shafer. "Propagating belief functions with local computations." *IEEE Expert* 1.3 (1986): 43-52.
[402] Shenoy, Prakash P., and Glenn Shafer. "Axioms for probability and belief-function propagation." *Machine intelligence and pattern recognition*. Vol. 9. North-Holland, 1990. 169-198.
[403] Jiang, Shengli, et al. "Uncertainty quantification for molecular property predictions with graph neural architecture search." *Digital Discovery* 3.8 (2024): 1534-1553.
[404] Liu, Jihao, et al. "Fnas: Uncertainty-aware fast neural architecture search." *arXiv preprint arXiv:2105.11694* (2021).
[405] Sclar, Melanie, et al. "Explore theory of mind: Program-guided adversarial data generation for theory of mind reasoning." *International Conference on Learning Representations*. Vol. 2025. 2025.
[406] Ji, Shihao, Zihui Song, and Jiajie Huang. "Credal Transformer: A Principled Approach for Quantifying and Mitigating Hallucinations in Large Language Models." *arXiv preprint arXiv:2510.12137* (2025).
[407] Mubashar, Muhammad, Shireen Kudukkil Manchingal, and Fabio Cuzzolin. "Random-set large language models." *arXiv preprint arXiv:2504.18085* (2025).
[408] Kim, Jungeum, Sean O'Hagan, and Veronika Rockova. "Adaptive uncertainty quantification for generative AI." *arXiv preprint arXiv:2408.08990* (2024).
[409] Kim, Jungeum, Sean O'Hagan, and Veronika Rockova. "Adaptive uncertainty quantification for generative AI." *arXiv preprint arXiv:2408.08990* (2024).
[410] Kudukkil Manchingal, Shireen, et al. "Random-set neural networks." *International Conference on Learning Representations*. Vol. 2025. 2025.
[411] Wang, Kaizheng, et al. "Credal deep ensembles for uncertainty quantification." *Advances in Neural Information Processing Systems* 37 (2024): 79540-79572.
[412] Yao, Kai, and Baoding Liu. "Uncertain regression analysis: An approach for imprecise observations." *Soft computing* 22.17 (2018): 5579-5582.
[413] Guo, Renkuan, Yanhong Cui, and Danni Guo. "Uncertainty linear regression models." *Journal of Uncertain Systems* 5.4 (2011): 286-304.
[414] Matthies, Hermann G., et al. "Parametric and uncertainty computations with tensor product representations." *IFIP Working Conference on Uncertainty Quantification*. Berlin, Heidelberg: Springer Berlin Heidelberg, 2011.
[415] Matta, David. "Strategic Theory of Mind (SToM): Extending game theory through cognitive modeling, strategic interaction, and the emergence of trust. Zenodo." 2026,
[416] Harré, Michael S. "What can game theory tell us about an AI 'theory of mind'?." *Games* 13.3 (2022): 46.
[417] Baker, Chris, Rebecca Saxe, and Joshua Tenenbaum. "Bayesian theory of mind: Modeling joint belief-desire attribution." *Proceedings of the annual meeting of the cognitive science society*. Vol. 33. No. 33. 2011.
[418] Stacy, Stephanie, et al. "A Bayesian theory of mind approach to modeling cooperation and communication." *Wiley Interdisciplinary Reviews: Computational Statistics* 16.1 (2024): e1631.
[419] Baker, Chris Lawrence. *Bayesian theory of mind: Modeling human reasoning about beliefs, desires, goals, and social relations*. Diss. Massachusetts Institute of Technology, 2012.
[420] Kleiman-Weiner, Max, et al. "Evolving general cooperation with a Bayesian theory of mind." *Proceedings of the National Academy of Sciences* 122.25 (2025): e2400993122.
[421] Westby, Samuel, and Christoph Riedl. "Collective intelligence in human-AI teams: A Bayesian theory of mind approach." *Proceedings of the AAAI Conference on Artificial Intelligence*. Vol. 37. No. 5. 2023.
[422] Teehan, Ryan, et al. "Emergent structures and training dynamics in large language models." *Proceedings of BigScience Episode# 5--Workshop on Challenges & Perspectives in Creating Large Language Models*. 2022.
[423] Berti, Leonardo, Flavio Giorgi, and Gjergji Kasneci. "Emergent abilities in large language models: A survey." *arXiv preprint arXiv:2503.05788* (2025).
[424] Schaeffer, Rylan, Brando Miranda, and Sanmi Koyejo. "Are emergent abilities of large language models a mirage?." *Advances in neural information processing systems* 36 (2023): 55565-55581.
[425] Sotomayor-Enriquez, Koraima, et al. "Open dataset of theory of mind reasoning in early to middle childhood." *Data in brief* 52 (2024): 109905.
[426] Poglitsch, Christian, et al. "Large language models for autism: evaluating theory of mind tasks in a gamified environment." *Scientific Reports* 15.1 (2025): 34763.

[427] Soubki, Adil, and Owen Rambow. "Machine theory of mind needs machine validation." *Findings of the Association for Computational Linguistics: ACL 2025*. 2025.
[428] Chandrasekaran, Arjun, et al. "It takes two to tango: Towards theory of AI's mind." *arXiv preprint arXiv:1704.00717* (2017).
[429] Kumar, Prabhat, Erin Zaroukian, and Douglas Summers-Stay. "Directions for Computational Theory of Mind: Data, Metrics, Models." *HCI International 2025–Late Breaking Papers: 27th International Conference on Human-Computer Interaction, HCII 2025, Gothenburg, Sweden, June 22–27, 2025, Proceedings, Part XV*. Springer Nature, 2026.
[430] Lu, Yi-Long, et al. "Do Theory of Mind Benchmarks Need Explicit Human-like Reasoning in Language Models?." *arXiv preprint arXiv:2504.01698* (2025).
[431] Kim, Hyunwoo, et al. "FANToM: A benchmark for stress-testing machine theory of mind in interactions." *Proceedings of the 2023 Conference on Empirical Methods in Natural Language Processing*. 2023.
[432] Riemer, Matthew, et al. "Position: Theory of mind benchmarks are broken for large language models." *arXiv preprint arXiv:2412.19726* (2024).
[433] Xu, Hainiu, et al. "OpenToM: A comprehensive benchmark for evaluating theory-of-mind reasoning capabilities of large language models." *Proceedings of the 62nd Annual Meeting of the Association for Computational Linguistics (Volume 1: Long Papers)*. 2024.
[434] Ma, Xiaomeng, Lingyu Gao, and Qihui Xu. "ToMChallenges: A principle-guided dataset and diverse evaluation tasks for exploring theory of mind." *Proceedings of the 27th Conference on Computational Natural Language Learning (CoNLL)*. 2023.
[435] Sclar, Melanie, et al. "Minding language models'(lack of) theory of mind: A plug-and-play multi-character belief tracker." *Proceedings of the 61st Annual Meeting of the Association for Computational Linguistics (Volume 1: Long Papers)*. 2023.
[436] Chen, Zhuang, et al. "ToMBench: Benchmarking theory of mind in large language models." *Proceedings of the 62nd Annual Meeting of the Association for Computational Linguistics (Volume 1: Long Papers)*. 2024.
[437] Sadhu, Jayanta, et al. "Multi-tom: Evaluating multilingual theory of mind capabilities in large language models." *arXiv preprint arXiv:2411.15999* (2024).
[438] Jones, Cameron R., Sean Trott, and Benjamin Bergen. "Comparing humans and large language models on an experimental protocol inventory for theory of mind evaluation (EPITOME)." *Transactions of the Association for Computational Linguistics* 12 (2024): 803-819.
[439] Mo, Zhenze, et al. "SUITE: Scaling Up Individualized Theory-of-Mind Evaluation in Large Language Models." *ToM4AI 2026* (2026): 77.
[440] Li, Mengfan, Xuanhua Shi, and Yang Deng. "RecToM: A Benchmark for Evaluating Machine Theory of Mind in LLM-based Conversational Recommender Systems." *Proceedings of the AAAI Conference on Artificial Intelligence*. Vol. 40. No. 37. 2026.
[441] Suzuki, Jundai, Ryoma Ishigaki, and Eisaku Maeda. "AnaToM: A Dataset Generation Framework for Evaluating Theory of Mind Reasoning Toward the Anatomy of Difficulty through Structurally Controlled Story Generation." *Proceedings of the 14th International Joint Conference on Natural Language Processing and the 4th Conference of the Asia-Pacific Chapter of the Association for Computational Linguistics*. 2025.
[442] Thiyagarajan, Prameshwar, et al. "Unitombench: Integrating perspective-taking to improve theory of mind in llms." *arXiv preprint arXiv:2506.09450* (2025).
[443] Yu, Fangxu, et al. "Persuasivetom: A benchmark for evaluating machine theory of mind in persuasive dialogues." *arXiv preprint arXiv:2502.21017* (2025).
[444] Yadav, Neemesh, et al. "DialToM: A Theory of Mind Benchmark for Forecasting State-Driven Dialogue Trajectories." *arXiv preprint arXiv:2604.20443* (2026).
[445] Zixiu Wu, Simone Balloccu, Vivek Kumar, Rim Helaoui, Ehud Reiter, Diego Reforgiato Recupero, and Daniele Riboni. 2022. Anno-MI: A Dataset of ExpertAnnotated Counselling Dialogues. In ICASSP 2022- 2022 IEEE International Conference on Acoustics, Speech and Signal Processing (ICASSP). 6177–6181. doi:10. 1109/ICASSP43922.2022.9746035
[446] Siyang Liu, Chujie Zheng, Orianna Demasi, Sahand Sabour, Yu Li, Zhou Yu, Yong Jiang, and Minlie Huang. 2021. Towards Emotional Support Dialog Systems. In Proceedings of the 59th Annual Meeting of the Association for Computational Linguistics and the 11th International Joint Conference on Natural Language Processing (Volume 1: Long Papers). 3469–3483.
[447] Xuewei Wang, Weiyan Shi, Richard Kim, Yoojung Oh, Sijia Yang, Jingwen Zhang, and Zhou Yu. 2019. Persuasion for Good: Towards a Personalized Persuasive Dialogue System for Social Good. In Proceedings of the 57th Annual Meeting of the Association for Computational Linguistics, Anna Korhonen, David Traum, and Lluís Màrquez (Eds.). Association for Computational Linguistics, Florence, Italy, 5635–5649. doi:10.18653/v1/P19-1566
[448] Zheng, Li, et al. "DebaToM: An Innovative Debate Framework for Theory of Mind Reasoning Using Large Language Models." *IEEE Transactions on Audio, Speech and Language Processing* (2026).
[449] Tong, Haibo, et al. "CogToM: A Comprehensive Theory of Mind Benchmark inspired by Human Cognition for Large Language Models." *arXiv preprint arXiv:2601.15628* (2026).
[450] Shinoda, Kazutoshi, et al. "Tomato: Verbalizing the mental states of role-playing llms for benchmarking theory of mind." *Proceedings of the AAAI Conference on Artificial Intelligence*. Vol. 39. No. 2. 2025.
[451] Le, Matthew, Y-Lan Boureau, and Maximilian Nickel. "Revisiting the evaluation of theory of mind through question answering." *Proceedings of the 2019 Conference on Empirical Methods in Natural Language Processing and the 9th International Joint Conference on Natural Language Processing (EMNLP-IJCNLP)*. 2019.
[452] Gandhi, Kanishk, et al. "Understanding social reasoning in language models with language models." *Advances in Neural Information Processing Systems* 36 (2023): 13518-13529.

[453] Gu, Yuling, et al. "Simpletom: Exposing the gap between explicit tom inference and implicit tom application in llms." *arXiv preprint arXiv:2410.13648* (2024).
[454] Kowalyshyn, Katharine, and Matthias Scheutz. "LLMs and their Limited Theory of Mind: Evaluating Mental State Annotations in Situated Dialogue." *arXiv preprint arXiv:2509.02292* (2025).
[455] Mao, Yuanyuan, et al. "BDIQA: A new dataset for video question answering to explore cognitive reasoning through theory of mind." *Proceedings of the AAAI Conference on Artificial Intelligence*. Vol. 38. No. 1. 2024.
[456] Li, Yuxuan, et al. "Egotom: Benchmarking theory of mind reasoning from egocentric videos." *arXiv preprint arXiv:2503.22152* (2025).
[457] Wei, Tingjiang, et al. "MovieGraph-ToM: Evaluating Long-Range Theory of Mind in Large Language Models via Implicit Social-Causal Graphs." *Proceedings of the AAAI Conference on Artificial Intelligence*. Vol. 40. No. 40. 2026.
[458] Tang, Yunlong, et al. "Video understanding with large language models: A survey." *IEEE Transactions on Circuits and Systems for Video Technology* (2025).
[459] Chen, Zhanwen, et al. "Through the theory of mind's eye: Reading minds with multimodal video large language models." *2025 International Joint Conference on Neural Networks (IJCNN)*. IEEE, 2025.
[460] Villa-Cueva, Emilio, et al. "MOMENTS: A Comprehensive Multimodal Benchmark for Theory of Mind." *arXiv preprint arXiv:2507.04415* (2025).
[461] Yin, Shukang, et al. "A survey on multimodal large language models." *National Science Review* 11.12 (2024): nwae403.
[462] Zhang, Duzhen, et al. "Mm-llms: Recent advances in multimodal large language models." *Findings of the Association for Computational Linguistics: ACL 2024* (2024): 12401-12430.
[463] Fan, Xianzhe, et al. "SoMi-ToM: Evaluating Multi-Perspective Theory of Mind in Embodied Social Interactions." *Advances in Neural Information Processing Systems* 38 (2026).
[464] Wilson, Jason R., et al. "ToMCAT: Benchmark for Socially Assistive Robots with Theory of Mind of Children Assembling Tangram Puzzles." *2025 34th IEEE International Conference on Robot and Human Interactive Communication (RO-MAN)*. IEEE, 2025.
[465] Zhang, Jian, et al. "Tangram puzzles in patients with neurocognitive disorders: a pilot study." *Psychiatry International* 4.4 (2023): 404-415.
[466] Shi, Haojun, et al. "Muma-tom: Multi-modal multi-agent theory of mind." *Proceedings of the AAAI Conference on Artificial Intelligence*. Vol. 39. No. 2. 2025.
[467] Bara, Cristian-Paul, CH-Wang Sky, and Joyce Chai. "MindCraft: Theory of mind modeling for situated dialogue in collaborative tasks." *Proceedings of the 2021 Conference on Empirical Methods in Natural Language Processing*. 2021.
[468] Melhart, David, Georgios N. Yannakakis, and Antonios Liapis. "I feel i feel you: A theory of mind experiment in games." *KI-Künstliche Intelligenz* 34.1 (2020): 45-55.
[469] Bortoletto, Matteo, Constantin Ruhdorfer, and Andreas Bulling. "ToM-SSI: Evaluating Theory of Mind in Situated Social Interactions." *Proceedings of the 2025 Conference on Empirical Methods in Natural Language Processing*. 2025.
[470] He, Liqi, et al. "Chain-of-Cognition Through Multi-Agent Collaboration Against Cross-Modal Cognitive Interference." *IEEE Transactions on Audio, Speech and Language Processing* (2026).
[471] Wang, Qiaosi, et al. "Towards mutual theory of mind in human-ai interaction: How language reflects what students perceive about a virtual teaching assistant." *Proceedings of the 2021 CHI conference on human factors in computing systems*. 2021.
[472] Soubki, Adil, et al. "Views are my own, but also yours: Benchmarking theory of mind using common ground." *Findings of the Association for Computational Linguistics: ACL 2024*. 2024.
[473] Magdalena Markowska, Mohammad Taghizadeh, Adil Soubki, Seyed Mirroshandel, and Owen Rambow. 2023. Finding common ground: Annotating and predicting common ground in spoken conversations. In Findings of the Association for Computational Linguistics: EMNLP 2023, pages 8221–8233, Singapore. Association for Computational Linguistics.
[474] Wilf, Alex, et al. "Think twice: Perspective-taking improves large language models' theory-of-mind capabilities." *Proceedings of the 62nd Annual Meeting of the Association for Computational Linguistics (Volume 1: Long Papers)*. 2024.
[475] Stagnitto, Serena Maria, et al. "TrAIngles: an LLM-Based Automatic Scoring Tool for Theory of Mind Assessments Across the Life Span 2." (2026).
[476] Liu, Yiwei, et al. "TactfulToM: Do LLMs Have the Theory of Mind Ability to Understand White Lies?." *Proceedings of the 2025 Conference on Empirical Methods in Natural Language Processing*. 2025.
[477] Mosqueira-Rey, Eduardo, et al. "Human-in-the-loop machine learning: a state of the art." *Artificial Intelligence Review* 56.4 (2023): 3005-3054.
[478] Michelson, Joel, et al. "Standoff: benchmarking representation learning for nonverbal theory of mind tasks." *2024 IEEE International Conference on Development and Learning (ICDL)*. IEEE, 2024.
[479] Sclar, Melanie, et al. "Explore theory of mind: Program-guided adversarial data generation for theory of mind reasoning." *International Conference on Learning Representations*. Vol. 2025. 2025.
[480] Lin, Zhiqiu, et al. "The clear benchmark: Continual learning on real-world imagery." *Thirty-fifth conference on neural information processing systems datasets and benchmarks track (round 2)*. 2021.
[481] Lomonaco, Vincenzo, and Davide Maltoni. "Core50: a new dataset and benchmark for continuous object recognition." *Conference on robot learning*. PMLR, 2017.
[482] Verwimp, Eli, et al. "Clad: A realistic continual learning benchmark for autonomous driving." *Neural Networks* 161 (2023): 659-669.

[483] Ståhl, Niclas, et al. "Evaluation of uncertainty quantification in deep learning." *International Conference on Information Processing and Management of Uncertainty in Knowledge-Based Systems*. Cham: Springer International Publishing, 2020.
[484] Klein, E.; Piot, B.; Geist, M.; Pietquin, O. A Cascaded Supervised Learning Approach to Inverse Reinforcement Learning. In *Proceedings of the Machine Learning and Knowledge Discovery in Databases, Prague, Czech Republic, 23–27 September 2013*; Lecture Notes in Computer Science; Blockeel, H., Kersting, K., Nijssen, S., Železný, F., Eds.; Springer: Berlin/Heidelberg, Germany, 2013; pp. 1–16
[485] David V. Pynadath and Stacy C. Marsella. 2005. PsychSim: Modeling Theory of MindwithDecision-Theoretic Agents. International Joint Conferences on Artificial Intelligence (2005), 1181–1186.
[486] Cuzzolin, Fabio, et al. "Knowing me, knowing you: theory of mind in AI." *Psychological medicine* 50.7 (2020): 1057-1061.
[487] Langley, Christelle, Fabio Cuzzolin, and Barbara J. Sahakian. "Neuroscience for AI: The importance of theory of mind." *Developments in Neuroethics and Bioethics*. Vol. 7. Academic Press, 2024. 65-83.
[488] Aru, Jaan, et al. "Mind the gap: Challenges of deep learning approaches to theory of mind." *arXiv preprint arXiv:2203.16540* (2022).
[489] Langley, Christelle, et al. "Theory of mind and preference learning at the interface of cognitive science, neuroscience, and AI: A review." *Frontiers in artificial intelligence* 5 (2022): 778852.